\documentclass[lettersize,journal]{IEEEtran}
\usepackage{cite}
\usepackage{graphicx}
\usepackage{caption}
\usepackage{multirow}
\usepackage{amsmath,amssymb,amsfonts}
\usepackage{amsthm}
\usepackage{mathrsfs}
\usepackage{textcomp}
\usepackage{manyfoot}
\usepackage{booktabs}
\usepackage{algorithm}
\usepackage{algorithmicx}
\usepackage{algpseudocode}
\usepackage{listings}
\usepackage{pgf-pie}
\usepackage{tikz}
\usepackage{float}
\usepackage{tabularx}
\usepackage{enumitem} 
\usepackage{lettrine} 
\usepackage{stfloats}
\usepackage{subcaption}
\usepackage{rotating} 
\usepackage{array} 
\usepackage{xcolor} 

\definecolor{darkgreen}{rgb}{0.0, 0.7, 0.0}  

\usepackage[colorlinks=true, linkcolor=blue, citecolor=darkgreen, urlcolor=blue]{hyperref}

\begin{document}

\title{EcoWeedNet: A Lightweight and Automated Weed Detection Method for Sustainable Next-Generation Agricultural Consumer Electronics}

\author{
    Omar H. Khater, Abdul Jabbar Siddiqui*, M. Shamim Hossain, Aiman El-Maleh%
    \thanks{O. H. Khater is with the Computer Engineering Department, King Fahd University of Petroleum and Minerals (KFUPM), Dhahran 31261, Saudi Arabia (e-mail: g202313250@kfupm.edu.sa).}%
    \thanks{*A. J. Siddiqui (Corresponding Author, Co-first author) is with the SDAIA-KFUPM Joint Research Center on Artificial Intelligence, IRC for Intelligent Secure Systems and Department of Computer Engineering, KFUPM (e-mail: abduljabbar.siddiqui@kfupm.edu.sa).}%
  \thanks{M. Shamim Hossain is with the Research Chair of Pervasive and Mobile Computing, and also with the Department of Software Engineering, College of Computer and Information Sciences, King Saud University, Riyadh 12372, Saudi Arabia (e-mail: mshossain@ksu.edu.sa)}
  \thanks{A. H. El-Maleh is the Chairman of the Department of Computer Engineering and a Professor at the College of Computing and Mathematics, King Fahd University of Petroleum \& Minerals, Dhahran 31261, Saudi Arabia (e-mail: aimane@kfupm.edu.sa)} 
}

\maketitle

\begin{abstract}

Sustainable agriculture plays a crucial role in ensuring world food security for consumers. 
A critical challenge faced by sustainable precision agriculture is weed growth, as weeds compete for essential resources with crops, such as water, soil nutrients, and sunlight, which notably affect crop yields. The adoption of automated computer vision technologies and ground agricultural consumer electronic vehicles in precision agriculture offers sustainable, low-carbon solutions. 
However, prior works suffer from issues such as low accuracy and precision, as well as high computational expense. This work proposes EcoWeedNet, a novel model that enhances weed detection performance without introducing significant computational complexity, aligning with the goals of low-carbon agricultural practices. The effectiveness of the proposed model is demonstrated through comprehensive experiments on the CottonWeedDet12 benchmark dataset, which reflects real-world scenarios. EcoWeedNet achieves performance comparable to that of large models (mAP@0.5 = 95.2\%), yet with significantly fewer parameters (approximately $\textbf{4.21\%}$ of the parameters of YOLOv4), lower computational complexity and better computational efficiency ($\textbf{6.59\%}$ of the GFLOPs of YOLOv4). These key findings indicate EcoWeedNet's deployability on low-power consumer hardware, lower energy consumption, and hence reduced carbon footprint, thereby emphasizing the application prospects of EcoWeedNet in next-generation sustainable agriculture. These findings provide the way forward for increased application of environmentally-friendly agricultural consumer technologies.

\end{abstract}

\begin{IEEEkeywords}
Weed Detection, energy-efficient consumer electronics, Parameter-Free Attention.
\end{IEEEkeywords}

\section{Introduction}

Modern agricultural consumer electronics are revolutionizing sustainable precision agriculture practices through the use of advanced tools and automated technologies \cite{AttiqueCE2024,YLi2024}
to enhance efficiency and lower the carbon footprint and environmental impact \cite{wu2022state}.
A critical challenge in sustainable agriculture is the pervasive problem of weed growth. Weeds compete with crops for vital resources such as nutrients, water, and sunlight, significantly reducing yields. 
Weed growth incurs a huge cost to the agriculture industry. For example, a report by Weed CRC mentions that the average net annual loss due to weed growth incurred by Australian agriculture is \$3.9 billion \cite{WeedCRC}. The report also estimates the cost of weed control to be at least \$19.6 million annually for weed control in national parks and natural environments.
Additionally, weeds are estimated to be presently responsible for huge crop losses in the agriculture industry \cite{RAZFAR2022100308}.
Globally, the impact is even more staggering, with weed management costs and losses reaching huge amounts.
Addressing the weed growth issue is pivotal for achieving sustainable agricultural practices and ensuring food security for a growing population \cite{car2023structured}.

\begin{figure}[h]
    \centering
    \includegraphics[width=\linewidth]{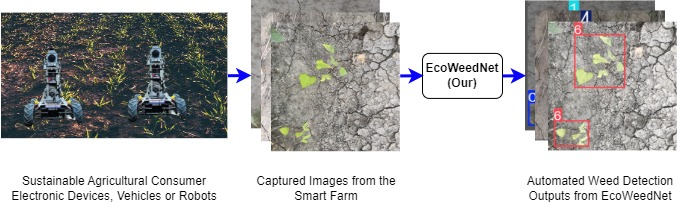}
    \caption{Illustration of the proposed lightweight, computationally- and energy-efficient EcoWeedNet model's performance in sustainable (low-carbon) consumer electronics-based precision agriculture applications for automated inspection and detection of weeds. }
    \label{fig:UGV_weed_detection}
\end{figure}

Even with recent significant advancements, weed detection in sustainable agriculture is challenging due to high computational needs that exceed the capabilities of consumer edge devices, energy inefficiency, changing accuracy with changes in field conditions, and dataset limitations. Additionally, hardware and software integration on agricultural consumer electronics and environmental factors affecting device reliability present challenges to real-time weed detection.

Weeds can be defined as undesirable plants that grow with other desirable plants or crops and compete with the crops for sunlight, water, and soil nutrients, and may host multiple pathogens. 
Intelligent consumer electronic aerial and ground vehicles have gained attention for automating the tasks of weed detection and control \cite{wu2023guest}\cite{nagarajan2024artificial}. This has been fueled by the recent advances in vision-based deep learning models that have been utilized to offer sustainable solutions for weed detection \cite{wang2019review}.

As illustrated in Figure~\ref{fig:UGV_weed_detection}. Vision-based models deployed on consumer electronic (CE) unmanned ground vehicles (UGVs) capture high-resolution images of the farmland \cite{al2023federated}, which are then processed by deep learning models such as convolutional neural networks (CNNs) to perform the task of weed detection. 
The problem of weed detection in images can be formulated as object detection. As such, there are broadly two categories of vision-based object detection methods: two-stage and one-stage detectors \cite{yang2022focal}. The former includes models such as Faster R-CNN, which are accurate but slow. The latter includes models such as the YOLO  family, which are relatively more computationally efficient, have lower energy consumption, and rapid inference capabilities, which can be beneficial in real-time low-carbon consumer agricultural applications \cite{du2024ai}.

Recently, there was a revolution in the deep learning community by the concept of attention in \cite{vaswani2017attention}, enabling the models to dynamically focus on the most relevant features in the input data, which positively affects the performance of the models across various tasks. The idea of attention is computing the weighted sum for each feature, and these weights help to give importance to each input feature, which allows the model to focus on the features that have high significance regarding the required task \cite{zhang2024aganet}.

Basically, the traditional attention modules are divided into spatial and channel attention \cite{song2024learning}. Channel attention enhances deep learning models by emphasizing the most informative features of inter-channel relationships. Conversely, spatial attention emphasizes inter-spatial interactions and accentuates the significant locations within the feature maps. Integrating spatial and channel attention will enable models to emphasize critical information, hence enhancing the performance of deep learning models in tasks such as object detection. The traditional attention modules add parameters that add more complexity and computation cost. To overcome this problem, we proposed to investigate the impact of the parameter-free attention modules on the deep learning models' performance, which relies on computing statistical measures to determine the importance of the features and maintain the model's computational efficiency.

In this paper, we make the following contributions:

\begin{itemize}[leftmargin=*]
    \item We design and develop a novel method for lightweight and automated weed detection by consumer electronic agricultural vehicles and robots for sustainable precision agriculture
  \item 
  The performance of our proposed model is investigated using a real-world dataset. Results bring new insights and demonstrate superior performance of the proposed method, proving the capability of our proposed model to be effective for the deployment on consumer electronic agricultural vehicles and robots. 


  \item 
   We demonstrate that the proposed model enhancements offer outstanding accuracy in detection tasks while maintaining the model's computational efficiency for  deployment on sustainable consumer electronic devices.

  
\end{itemize}


\section{Related Works}

Generally, object detectors are divided into one-stage detectors and two-stage detectors, highlighting the trade-off between high performance and computational complexity. Additionally, the parameter-free attention modules play a vital role in enhancing the model performance without adding any parameters or FLOPs, making the model suitable for deployment on consumer electronic devices.

\subsection{Object Detectors: Two-Stage vs. One-Stage Approaches}

Object detection methods for weed detection in precision agriculture can be generally divided into two groups, including two-stage and one-stage detectors, both having weaknesses and strengths.

The two-stage detectors, such as Faster R-CNN, rely on region propositions and then classify. In \cite{rahman2023performance}, Faster R-CNN tested CottonWeedDet3 with 848 RGB images for three weed categories. As effective in defining fine-grain details in complex backgrounds, the model struggled to detect smaller objects. In a move to counter such vulnerability, \cite{fan2023deep} proposed additions such as a Convolutional Block Attention Module (CBAM) for enhancing informative features, a Bidirectional Feature Pyramid Network (BiFPN), and a Bilinear Interpolation algorithm for multi-scale weed detection. With such, mAP rose from 79.98\% to 98.43\%, but performance lowered to 92.81\% in the case of nightfall, highlighting a two-stage model vulnerability in heterogeneous environments.

On the one hand, one-stage detectors such as the YOLO family produce end-to-end processing with efficient inference and, therefore, can work for real-time cases. \cite{rahman2023performance} contrasted YOLOv5n with Faster R-CNN with CottonWeedDet3 dataset and exhibited YOLOv5n with a record 17 ms inference, outpacing Faster R-CNN in terms of efficiency but with competitive accuracy. In contrast, one-stage architectures lack small-object detection in comparison with two-stage detectors, for which region proposals work in their favour.

Some one-stage detector optimizations have resolved such problems. \cite{fan2024yolo} optimized YOLOv5n with integration with backbone with ShuffleNet-v2, integration with a parallel hybrid attention module, and three-level BiFPN for feature fusion improvement in it. By such optimizations, model complexity was reduced by over 80\%, with an inference time of 12 ms less and mAP@0.5 accuracy of 97.8\%, while the model still struggles under severe occlusions. Likewise, \cite{rai2024agricultural} optimized YOLOv7 with YOLO-Spot\_M through pruning and architectural optimizations. In it, 75.3\% and 82.4\% of its parameters and computational cost, respectively, reduced and exhibited 5 times improvement in terms of processing pace at a mAP@0.5 accuracy of 80.6\%, with suitability for a run in a less-resourced environment such as Jetson AGX Xavier. On the other hand, the generalization is restricted because the dataset includes only four weed species.

From a general consideration, even two-stage detectors outperform in terms of accuracy and processing of small but with high computational expense, and therefore, hardly work practically for real-time cases in practice. On the one hand, one-stage detectors have a balanced trade-off with a high inference pace and, therefore, can work effectively in a less-latency, less-resource environment, according to \cite{hu2024real}.

\begin{table*}[t]
    \centering
    \caption{Summary of Related Works and their Limitations }%
    
    \resizebox{\textwidth}{!}{%
    \begin{tabular}{c|c|c}
        \hline
        \textbf{Reference} & \textbf{Method} & \textbf{Limitations} \\
        \hline
        \cite{rahman2023performance} & Faster R-CNN on CottonWeedDet3 & Struggles with small object detection in complex backgrounds. \\
        \hline
        \cite{fan2023deep} & Faster R-CNN + CBAM, BiFPN, Bilinear Interpolation & Performance drops under nighttime conditions. \\
        \hline
        \cite{rahman2023performance} & YOLOv5n vs. Faster R-CNN & Efficient (17 ms inference) but weaker small-object detection. \\
        \hline
        \cite{fan2024yolo} & YOLOv5n + ShuffleNet-v2 + PHAM + BiFPN & Struggles under severe occlusions. \\
        \hline
        \cite{rai2024agricultural} & YOLOv7 with pruning (YOLO-Spot\_M) & Limited generalization due to dataset (only 4 weed species). \\
        \hline
        \cite{guo2024lightweight} & SERMAttention + BiFPN & Dataset lacks diversity, limiting real-world application. \\
        \hline
        \cite{diao2023spatial} & Res-3D-OctConv (Spatial and Spectral Attention) & PCA-based feature selection may lose crucial details. \\
        \hline
        \cite{karim2024development} & CBAM + YOLOv8n (YOLOv8n-CBAM-C3Ghost) & Dataset imbalance affects performance across weed classes. \\
        \hline
        \cite{fan2024yolo} & PHAM + BiFPN in YOLO-WDNet & Increases model complexity and computational demand. \\
        \hline
        \cite{hu2024real} & ECA + CA in YOLOv7-L & Additional complexity (DownC, ELAN-B3) affects suitability for edge devices. \\
        \hline
    \end{tabular}%
    }
    \label{tab:related_works}
\end{table*}

\subsection{Attention Mechanisms}

Nowadays, attention mechanisms have become indispensable in enhancing the performance of deep learning models, especially for tasks that require precise feature extraction and complex backgrounds. By focusing on the spatial and channel features, the attention modules assist the deep learning models in becoming more robust and reliable across several applications. 
There are two broad categories of attention mechanisms: (i) parameter-based and (ii) parameter-free. 
In the parameter-based attention mechanisms, the channel and spatial attention modules depend on extracting the informative features using convolutional layers, which add more parameters and floating point operations (FLOPs). In contrast, the parameter-free attention modules are engineered to improve the model performance by relying on operations to compute the importance of the features without adding parameters or FLOPs to the architecture.

The SERMAttention module is proposed in \cite{guo2024lightweight}, which is considered a channel attention mechanism to emphasize the highest informative channels in the feature maps. Consequently, the model performance improved in terms of weed identification. This mechanism integrated with BiFPN to improve the multi-scale feature fusion, which led to a lightweight model suitable for computational-constrained devices. Despite the advantages of the proposed attention module, the paper showed some limitations regarding the complex real-world environment because of the lack of diversity in the weed classes. Based on that, the diversity in the dataset is vital to utilize the potential of the SERMAttention module in real-world applications. On the other hand, the spatial attention modules emphasize the spatial regions in the feature maps. 

In \cite{diao2023spatial}, the authors proposed that the Res-3D-OctConv framework integrates spatial and spectral attention to enhance the hyperspectral imaging models for weed detection. As a result, the model achieved notable improvements, scoring an accuracy of 98.56\% and outperforming the Support Vector Machine (SVM) and K-Nearest Neighbour (KNN) with 10.20\% and 8.65\%, respectively. The model relied on Principal Component Analysis (PCA), which rarely loses crucial information during feature aggregation. In \cite{woo2018cbam}, the writers proposed an enhanced attention module that integrated spatial and channel attention in one block. The Convolutional Block Attention Module (CBAM) refines the feature maps at different levels. The enhanced attention block is validated across multiple CNN architectures, such as ResNet and MobileNet, without notably increasing the computational cost. While, CBAM is implemented sequentially, which can increase the computational latency compared with the parallel attention mechanism. Similarly, the authors in \cite{karim2024development} investigated more CBAM to improve the performance of the detection models while maintaining lightweight. The integration of the CBAM to YOLOv8n to create the YOLOv8n-CBAM-C3Ghost model allowed more focus on the channel and spatial information in the input tensor, achieving a mAP@0.5 of 97.6\%, while the model's number of parameters is 3.61 million. The used dataset suffers from unbalancing across some weed classes.

Recently, the growing trend has highlighted attention modules that are parameter-free and do not add any parameters or FLOPs to the architecture, which overcomes the constraint of installing the traditional attention mechanism based on the convolutional layers that add parameters to the models and make the computations more costly. The SimAM module is proposed in \cite{yang2021simam}, which is inspired by neuroscience and can be one of the most efficient parameter-free attention modules nowadays by refining the features at a neuron level without adding any trainable parameters. The writers validated the proposed parameter-free attention module by installing the block to ResNet-18, and they noticed a significant improvement in top-1 accuracy on ImageNet without adding any parameter or FLOP. Although the SimAM is a parameter-free attention module, it adds a little latency in the inference. Similarly, the Swift Parameter-Free Attention Block (SPAB) is introduced in \cite{Wan_2024_CVPR} for single-image super-resolution tasks. The proposed parameter-free attention block utilized residual connections and symmetric activation functions to enhance the feature representation while expressing the redundant information. The SPAB outperformed parameter-heavy models, such as IMDN and ShuffleMixer, despite its simplicity. The significant improvements that SPAB adds to the vision-based models make it a strongly competitive option between the parameter-free attention modules.

The authors in \cite{fan2024yolo} proposed a hybrid attention module called parallel hybrid attention mechanism (PHAM). The proposed module tries to balance the strengths of channel and spatial attention. By combining the PHAM with a three-level BiFPN, a significant reduction in terms of complexity while maintaining the high performance of the proposed YOLO-WDNet model, and achieved a mAP@0.5 of 97.8\%. The attention mechanism combination helped the model achieve multi-scale feature fusion, which can be considered a vital requirement for weed detection tasks, but PHAM makes the model more complex and demands extra computation.

Guided Supervised Attention (GSA) is proposed in \cite{liu4903438gsa}, which generates spatial attention masks from objectness predictions without adding any learnable parameters. GSA offered an improvement regarding feature extraction in complex environments, illustrating its impact in challenging scenarios. GSA increases inference time by 10-15\%, which may impact real-time performance, especially in latency-sensitive applications. Similarly, in \cite{hu2024real}, the Efficient Channel Attention (ECA) and Coordinate Attention (CA) modules were installed into YOLOv7-L to enhance the spatial and channel feature extraction, which enhanced the accuracy of the detection in different conditions. However, these modules added notable improvement in the performance of the model; their reliance on DownC and ELAN-B3 added some complicity to the architecture and made this model less suitable for consumer electronic devices.

In conclusion, the attention mechanisms proved effective in the vision-based deep learning models, showing their ability to refine the features and improve the performance in terms of spatial and channel dimensions. To be more specific for edge device applications, the parameter-free attention modules combined both efficiency and accuracy and became the optimal choice for real-time applications.


\section{Proposed Method}

Considering the constraints of sustainable consumer electronic devices and precision low-carbon agriculture, this work proposes a novel lightweight and automated weed detection method. The enhanced model offers precise agriculture weed detection, exploring the role of two parameter-free attention modules: SPAB in the backbone and the so-called Simplified Attention Module, SimAM, both in the backbone and neck. The modules reinforce the model by paying more attention to informative features while maintaining computational efficiency, which is crucial for real-time applications. SPAB refines feature extraction, while SimAM guarantees overall refinement of the features across successive stages of processing the variability of an agricultural environment. Therefore, our approach paves the way for adopting intelligent and energy-efficient agriculture solutions that balance high performance with minimum harmful environmental impact, thus fully fitting with sustainable agricultural practices.

\begin{figure*}[!t]
\centering
\fbox{\includegraphics[width=\textwidth, keepaspectratio]{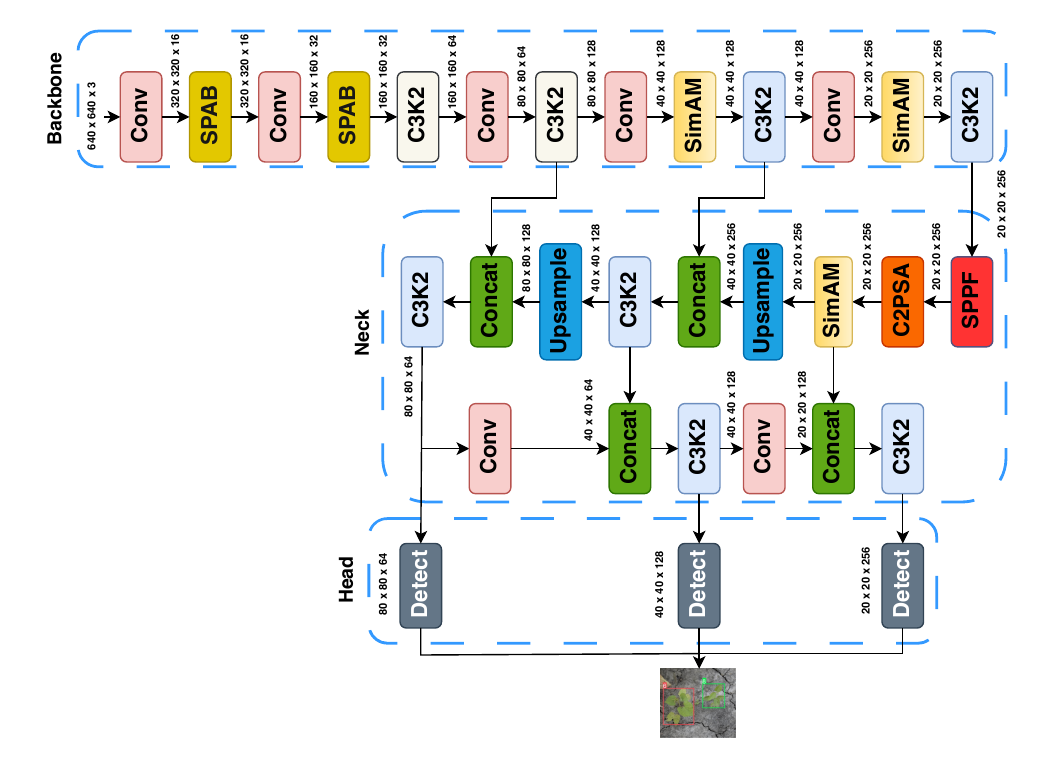}}
\caption{Architecture of the proposed EcoWeedNet model.}
\label{fig:EcoWeedNet}
\end{figure*}

\begin{figure*}[!t]
\centering
\fbox{\includegraphics[width=\textwidth, keepaspectratio]{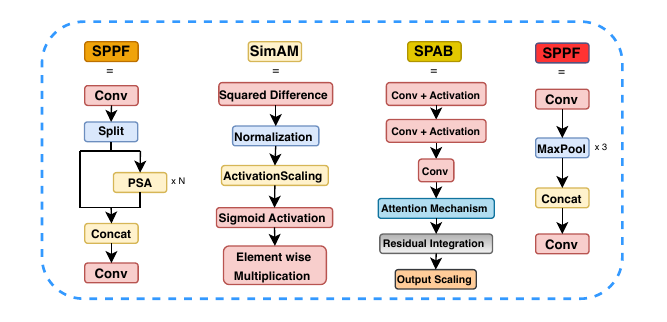}}
\caption{Important blocks in the architecture of the proposed EcoWeedNet model.}
\label{fig:EcoWeedNet_blocks}
\end{figure*}

This section provides a detailed overview of the EcoWeedNet architecture, as shown in Figure \ref{fig:EcoWeedNet}, highlighting its three key components: the backbone, neck, and head. Additionally, it explores the integration and functionality of two state-of-the-art attention modules, SimAM and SPAB, within the system, and the important blocks are explained in Figure \ref{fig:EcoWeedNet_blocks}. Additionally, the abbreviations of the blocks used in the EcoWeedNet architecture are listed in Table~\ref{tab:symbols_abbreviations}.

\subsection{Backbone}

The backbone of the EcoWeedNet model contains multiple blocks that effectively impact the feature extraction process, highlighting the most informative parameters in the feature map. Convolutional layers contribute to identifying the patterns that are crucial for the detection tasks. Additionally, C3K2 is a sophisticated block designed for better feature extraction. C3K2 block utilizes dual convolutional operations and can also notably lower the computational load by integrating more C3K blocks to reduce the spatial dimensions while making the feature map deeper.

Moreover, the Swift Parameter-Free Attention Block (SPAB) is integrated to enhance the model's focus on informative features, boosting detection accuracy while maintaining the model is lightweight. The simple attention module (SimAM) inspired by neuroscience enhances the feature map representations without adding learnable parameters and floating point operations (FLOPs).

These modules have a significant impact on the enhancements of the feature extraction. These modules optimize the enhanced model for high performance and computational efficiency, making it an optimal choice for edge device deployment. The enhancements ensure reliable weed detection and provide precise and quick performance.

\subsection{Neck}

The neck has an essential role in enhancing and synthesizing the feature maps from the backbone and passing them to the detection heads. This component involves multiple modules.

Spatial Pyramid Pooling - Fast (SPPF) captures multi-scale contextual information from the input tensor. It contains convolutional layers to reduce the spatial dimensions, followed by max pooling at multiple scales. Varied receptive fields are generated, which are crucial for detecting varied sizes of weeds. Then, the features are concatenated, ensuring the rich representation of input data.

Convolutional Block with Parallel Spatial Attention (C2PSA) focuses on the spatial patterns in the feature maps by utilizing a split-transform-merge strategy, where the input features are split, transformed using self-attention blocks (PSABlocks), and then merged back. This module has a notable impact on emphasizing the most informative regions in the feature maps while expressing the less important features, contributing positively to the weed detection task.

The importance of convolutional and C3K2 layers is still essential in the neck, preparing the data for further refinement and ensuring the important feature extraction for accurate weed detection. Moreover, the SimAM was installed in the neck before the first upsampling to ensure that the only relevant and informative feature would be scaled up, which enhances the overall quality of the spatial information processing in subsequent layers.

Upsampling is utilized to scale up the spatial dimensions of the input feature maps, restoring the details that were compressed in the earlier processes. The concatenation contributes to enriching the feature maps with suppressed information to boost the ability of the model to offer robust performance.

Linked through modules, all these guarantees that feature extraction and processing are effectively done in the enhanced model architecture, dynamically focusing and refining them for further boosts in general detection performance, becoming especially suitable for real-time applications on consumer electronic devices.

\subsection{Head}

In EcoWeedNet, a head generates a prediction for classification and localization with a multi-scale scheme for detection. For improved weed detection for a diversity of scales, three output heads at three scales, namely 80 × 80 for small weeds, 40 × 40 for weeds with a medium one, and 20 × 20 for weeds with a larger one, are utilized. With a multi-scale scheme, accuracy in detection is increased, and robust performance in a diversity of agricultural scenarios, even in occluded and occluded-overlapping ones, is attained. With such a scheme, effective and efficient weed detection is acquired through EcoWeedNet, and it can run in real-time on consumer electronic devices in low-carbon and environmentally friendly agriculture.

\vspace{-0.5cm}

\subsection{SimAM}

SimAM is a parameter-free attention mechanism designed to enhance feature map representations \cite{yang2021simam}. SimAM directly computes 3D attention weights without adding learnable parameters or increasing FLOPs, making it more efficient than traditional attention modules, offering high performance without computational complexity, outperforming the traditional attention mechanism, which relies on the convolutional layers to enhance the model performance and adds computational complexity.

The key principle of the SimAM is designing the energy function inspired by neuroscience, which is utilized to evaluate the importance of each neuron in the feature map. The energy function emphasizes the response of each neuron compared to its surrounding neurons. SimAM aims to highlight neurons with low energy from their surrounding neurons to evaluate their importance. The energy function \( e_t(w_t, b_t, y, X) \) is defined as:

    \begin{align}
e_t(w_t, b_t, y, X) = &\ \frac{1}{\begin{array}{c} M \\ -1 \end{array}} \sum_{i=1}^{M-1} \left(-1 - (w_t x_i + b_t)\right)^2 \nonumber \\
&+ \left(1 - (w_t t + b_t)\right)^2 + \lambda w_t^2
\label{eq:energy_function}
\end{align}

where: $t$: Target neuron's value, $x_i$: Surrounding neurons' values, $M$: Total number of neurons in a channel, $w_t$, $b_t$: Linear transform parameters (weight and bias), $\lambda$: Regularization parameter.

The neurons with low energy are highlighted with high distinctiveness and considered more informative for feature representation. To find the optimal weights, we derive a closed-form solution for $w_t$ and $b_t$:

    \begin{equation}
w_t = -\frac{2(t - \mu_t)}{(t - \mu_t)^2 + 2\sigma_t^2 + 2\lambda}
\label{eq:optimal_weight}
\end{equation}

\begin{equation}
b_t = -\frac{1}{2}(t + \mu_t)w_t
\label{eq:optimal_bias}
\end{equation}

Where:
\begin{equation}
\mu_t = \frac{1}{M - 1} \sum_{i=1}^{M-1} x_i, \quad \sigma_t^2 = \frac{1}{M - 1} \sum_{i=1}^{M-1} (x_i - \mu_t)^2
\label{eq:mean_variance}
\end{equation}

Here, \( \mu_t \) is the mean, and \( \sigma_t^2 \) is the variance of the surrounding neurons.

The SimAM mechanism aims to minimize the energy, and the minimal energy for the target neuron \( t \) is given by:

\begin{equation}
e^*_t = \frac{4(\sigma_t^2 + \lambda)}{(t - \mu_t)^2 + 2\sigma_t^2 + 2\lambda}
\label{eq:min_energy}
\end{equation}

The importance of the neuron is then inversely proportional to this minimal energy:

    \begin{equation}
\text{Importance} = \frac{1}{e^*_t}
\label{eq:importance}
\end{equation}

The SimAM is characterized by fast and efficient computation of neuron importance by utilizing simple statistical measures, such as the mean and variance of the feature map, thanks to the closed-form solution for the energy function.

On the other hand, traditional attention mechanisms operate on spatial or channel dimensions, which adds complexity to the model. Computing comprehensive 3D attention weights captures spatial and channel-wise interactions simultaneously, enhancing the feature refinement operations. The refined feature map $X'$ is calculated as:

    \begin{equation}
X' = \sigma\left(\frac{1}{E}\right) \odot X
\label{eq:refined_feature_map}
\end{equation}

Where \( E \) contains the minimal energies \( e^*_t \) for all neurons in the feature map, and \( \sigma(\cdot) \) is the sigmoid function to ensure the scaling factor lies within \( [0, 1] \). The symbol \( \odot \) denotes element-wise multiplication.

SimAM does not offer any learnable parameters or complexity to the model and is considered a plug-and-play module that can be integrated easily into any model network, improving the performance with computational complexity.

\subsection{SPAB}

The Swift Parameter-Free Attention Block (SPAB) is designed to be lightweight (hence suitable for consumer electronic devices) and enhance the performance of the models by highlighting the most informative parameters in the feature maps \cite{Wan_2024_CVPR}. 

SPAB relies on an effective structure consisting of three convolutional layers, which capture the crucial patterns in the feature map, such as textures and edges. The output of the $j$-th convolutional layer is denoted as $\mathbf{H}_i$, where each layer applies a kernel $\mathbf{W}_i$ to the input from the previous layer and the convolution operation is represented as:

    \begin{equation}
\mathbf{H}_i = \mathbf{F}_i \left( \mathbf{W}_i \ast \mathbf{O}_{i-1} \right),
\label{eq:conv_operation}
\end{equation}

where $\ast$ represents the convolution operation and $\mathbf{O}_{i-1}$ is the output of the previous layer.

Then, a symmetric activation function $\sigma_a$ is applied to create an attention map to emphasize the informative regions. The generated attention map contains the original features while highlighting the rich regions and depressing the less important or redundant features. This attention map $\mathbf{V}_i$ is calculated as:

    \begin{equation}
\mathbf{V}_i = \sigma_a \left( \mathbf{H}_i \right),
\label{eq:attention_map}
\end{equation}

Where $\sigma_a$ is a symmetric activation function, typically chosen to be a variant of the sigmoid function, which ensures that the activation is balanced around the origin. Moreover, the SPAB structure addressed the information loss by utilizing residual connections to ensure that there was no loss in the input data during the processing operations. Overall, the SPAB architecture provides a robust enhancement and improved feature extraction while maintaining the model's lightweight and suitability for resource-constrained devices.

The generated attention map contains the original features while emphasizing the rich regions and depressing the less important or redundant features. The final output of the SPAB block is obtained by performing an element-wise multiplication of the original feature map $\mathbf{U}_i$ and the attention map $\mathbf{V}_i$:

    \begin{equation}
\mathbf{O}_i = \mathbf{U}_i \odot \mathbf{V}_i,
\label{eq:output_feature}
\end{equation}

Where $\odot$ represents element-wise multiplication, and $\mathbf{U}_i$ is the feature map obtained by adding residual connections from the input:

    \begin{equation}
\mathbf{U}_i = \mathbf{O}_{i-1} \oplus \mathbf{H}_i,
\label{eq:residual_connection}
\end{equation}

Where $\oplus$ denotes element-wise summation.

Existing weed detection algorithms in agricultural consumer electronics usually suffer from high computational complexity, energy inefficiency, poor detection accuracy in diverse field conditions, poor generalization with class imbalance, and poor integration on embedded platforms. EcoWeedNet addresses some of these limitations by introducing parameter-free attention modules to a baseline architecture. This thus significantly enhances the detection accuracy and improves the weed detection in diverse environments while maintaining the computational complexity roughly the same, making it highly suitable for real-time deployment on resource-limited edge hardware.

\begin{table}[h]
\centering
\caption{
{%
  \parbox{0.66\linewidth}{%
    \centering
    Abbreviations used in Figure~\ref{fig:EcoWeedNet}.
  }%
}
}
\label{tab:symbols_abbreviations}
\renewcommand{\arraystretch}{1.6}
\resizebox{\columnwidth}{!}{%
\begin{tabular}{>{\centering\arraybackslash}m{4cm}|>{\centering\arraybackslash}m{10cm}}
\hline
\textbf{Symbol/Abbreviation} & \textbf{Meaning} \\ \hline
Conv     & Convolutional Layer \\
SPAB     & Swift Parameter-Free Attention Block \\
SimAM    & Simplified Attention Module \\
SPPF     & Spatial Pyramid Pooling–Fast \\
C2PSA    & Convolutional Block with Parallel Spatial Attention \\
C3K2     & Cross-stage partial connections block (CSP) with 3 convolutions and 2 residual connections \\
Detect   & Detection Head \\
Concat   & Concatenation Operation \\
Upsample & Increasing spatial resolution of feature maps \\
Backbone & Initial feature extraction part of the network \\
Neck     & Feature enhancement and aggregation stage \\
Head     & Final detection and classification stage \\ \hline
\end{tabular}%
}
\end{table}

\section{Experimental Setup}

\subsection{Dataset}
\begin{figure}
    \centering
    \setlength{\fboxsep}{0pt} 
    \setlength{\fboxrule}{1pt} 
    \fbox{\includegraphics[width=\columnwidth, keepaspectratio]{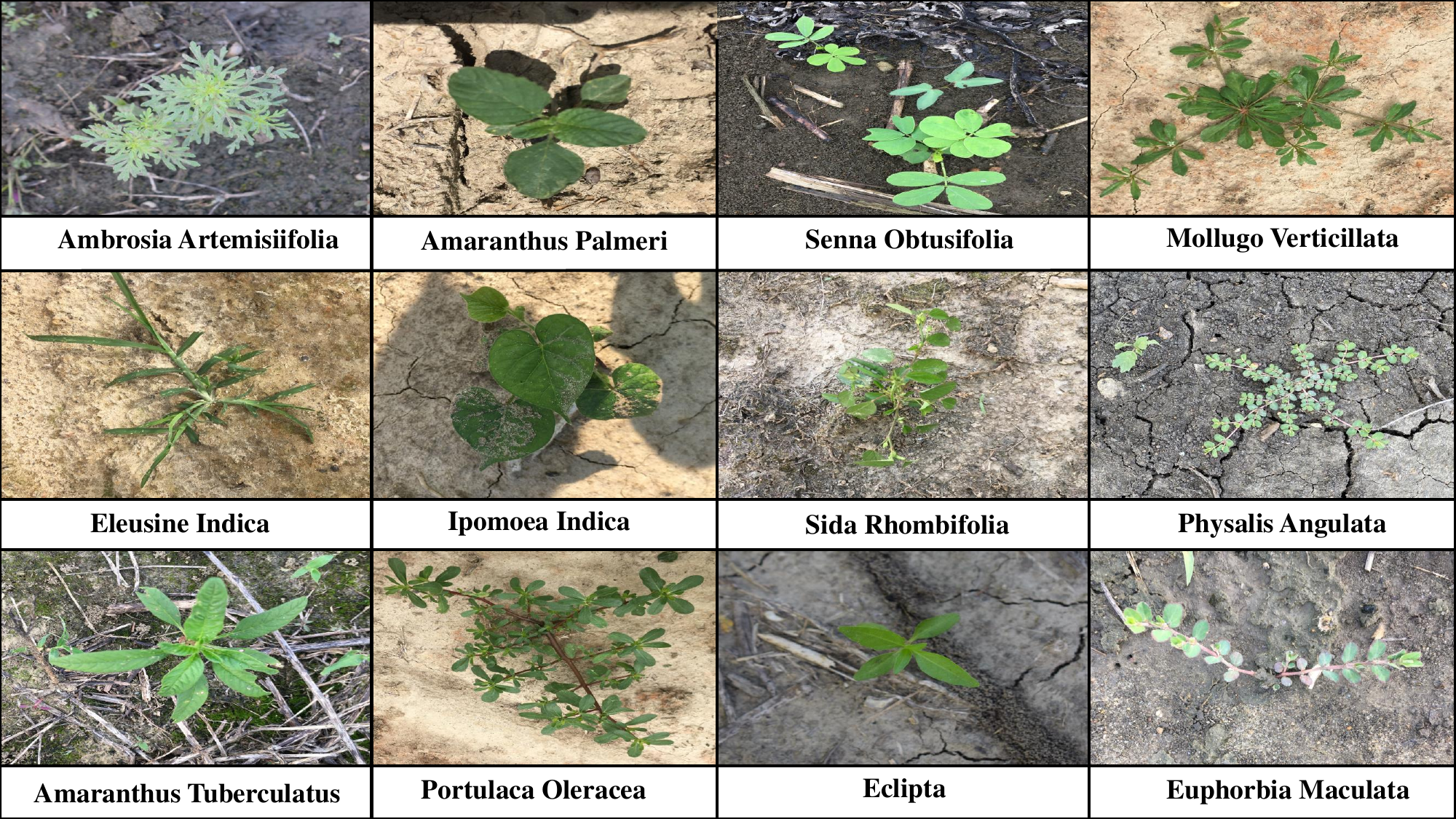}}
    \caption{Dataset Sample}
    \label{fig:dataset_samples}
\end{figure}

In our work, we used a CottonWeedDet12 dataset that consists of 12 common weed species in cotton fields in the southern region of the United States of America \cite{lu2023cottonweeddet12}, providing a strong and varied testing of EcoWeedNet. The dataset contains 5,648 high-resolution RGB images annotated with 9,370 bounding boxes, making CottonWeedDet12 a perfect openly available benchmark, as shown in Table~\ref{tab:dataset_statistics}. The used dataset includes the shadow effect, complex background, and multiple weed kinds per image, as shown in Figure~\ref{fig:dataset_samples}, emphasizing the real-world scenarios and enhancing the model performance, robustness, and generalization.

\begin{table}[H]
\centering
\caption{Weed Detection Dataset Statistics. The number of bounding boxes represents the total number of annotated weed instances.}
\resizebox{\columnwidth}{!}{%
\begin{tabular}{ccc}
\hline
\textbf{Weed Type}           & \textbf{Number of Images} & \textbf{Number of Bounding Boxes} \\ \hline
Eclipta & 576   & 865 \\ 
Ipomoea indica  & 1,149   & 1,344 \\ 
Eleusine indica  & 183    & 214  \\ 
Sida rhombifolia  & 433  & 486   \\ 
Physalis angulata & 111   & 123 \\ 
Senna obtusifolia   & 198   & 243  \\ 
Amaranthus palmeri  & 305   & 348  \\ 
Euphorbia maculata  & 671 & 956  \\ 
Portulaca oleracea  & 652 & 992 \\ 
Mollugo verticillata  & 564 & 962 \\ 
Amaranthus tuberculatus & 1,413 & 1,959  \\ 
Ambrosia artemisiifolia & 512 & 896  \\ \hline
\end{tabular}%
}
\label{tab:dataset_statistics}
\end{table}

It was divided into 80\% for training, 10\% for validation, and 10\% for testing.

Images were captured under diverse environmental settings. This does contribute significantly to the generalization of the model to real-world scenarios. This rich dataset forms a robust backbone for training and testing the proposed model to perform efficiently in weed detection tasks.

\subsection{Hardware and Software}

Through our experiments, we used an NVIDIA GeForce RTX 3080 Ti GPU (12 GB RAM), which offers high performance and is suitable for training our proposed model. We imported the baseline YOLOv11 architecture from the Ultralytics library. Moreover, we utilized the PyTorch framework, which offers high flexibility.

\subsection{Evaluation Metrics}

The evaluation metrics used to evaluate the model’s performance are presented below, in table \ref{tab:evaluation_metrics}:

\begin{table}[htbp]
\centering
\renewcommand{\arraystretch}{1.5} 
\caption{Evaluation metrics used for weed detection. Precision, Recall, and mean Average Precision (mAP) are calculated based on True Positives (TP), False Positives (FP), and False Negatives (FN).}

\label{tab:evaluation_metrics}
\begin{tabular}{ccm{5cm}} 
\hline
\textbf{Metric} & \textbf{Formula} & \textbf{Description} \\ \hline
Precision & $\displaystyle \frac{\text{TP}}{\text{TP} + \text{FP}}$ & Measures the ratio of correctly identified instances to all positive predictions. \\
Recall & $\displaystyle \frac{\text{TP}}{\text{TP} + \text{FN}}$ & Represents the model's ability to detect all actual positive instances. \\ 
mAP & $\displaystyle \frac{1}{N} \sum_{i=1}^{N} \text{AP}_i$ & Evaluates the average precision across all classes, ensuring reliable performance monitoring. \\ \hline
\end{tabular}
\end{table}


\begin{table*}[t]
\centering
\caption{Comparative Performance of the Proposed Model}
\resizebox{\textwidth}{!}{%
\begin{tabular}{>{\centering\arraybackslash}m{1.5cm}cccccccc}
\hline
\textbf{Model} & \textbf{SPAB Index} & \textbf{SimAM Index} & \textbf{Precision (\%)} & \textbf{Recall (\%)} & \textbf{mAP50 (\%)} & \textbf{mAP(50-95) (\%)} & \textbf{Parameters} & \textbf{(GFLOPs)} \\

\hline
 & 3 & 1,10 & 91.4 & 88.1 & 94.2 & 87.4 & 2.74M & 7.9 \\ 
 & 1 & 3,10 & 90.2 & 88.1 & 94.8 & 88.7 & 2.63M & 7.9 \\ 
 & 1 & 3,6,9,12 & 91.4 & 90.2 & 94.7 & 88.6 & 2.63M & 7.9 \\
 & 1 & 3,6,9 & 92.6 & 87 & 94.1 & 87.2 & 2.63M & 7.9 \\
 & 1 & 5,8,11,15,19 & 93.5 & 88.1 & 95.0 & 88.6 & 2.63M & 7.9 \\ 
 & 1 & 5,8,11,15,19,24,28 & 92.4 & 89.1 & 94.6 & 88.5 & 2.63M & 7.9 \\
 & 1 & 7,10,14,18 & 93.6 & 87.2 & 94 & 87.3 & 2.63M & 7.9 \\
{\textbf{Proposed}} & 1,3 & 8,11,15,19 & 92 & 89.4 & 95.2 & 88.6 & 2.78M & 9.3 \\ 
 {\textbf{Model}} & \textcolor{blue}{\textbf{1,3}} & \textcolor{blue}{8,11,15} & \textcolor{blue}{\textbf{93.2}} & \textcolor{blue}{\textbf{89}} & \textcolor{blue}{\textbf{95.2}} & \textcolor{blue}{\textbf{88.9}} & \textcolor{blue}{\textbf{2.78M}} & \textcolor{blue}{\textbf{9.3}} \\
 & 1,3 & 8,11,18 & 91.7 & 89 & 94.6 & 87.8 & 2.78M & 9.3 \\ 
 & 1,3 & 8,11,15,19 & 90.2 & 89.5 & 94.1 & 88 & 2.78M & 9.3 \\ 
 & 18 & 1,3,10,14 & 90.2 & 89 & 94 & 87.7 & 5.05M & 7.9 \\ 
 & 14 & 1,3,10,18 & 90 & 88.9 & 94.5 & 87.5 & 12.4M & 7.9 \\
 & 1 & 3,10,21 & 91.1 & 88.1 & 93.9 & 87.9 & 2.63M & 7.9 \\
 & 1 & 9,20 & 94.3 & 88 & 94.3 & 88.5 & 2.78M & 9.3 \\ 
\hline
\textbf{YOLO11n\cite{ultralytics2024yolo11}} & -- & -- &  89 & 88.6 & 93 & 85.6 & 2.6 M &  6.5 \\ 
\hline
\textbf{YOLO12n\cite{ultralytics2025yolov12}} & -- & -- &  92.8 & 84 & 93.2 & 86.9 & 2.6 M &  6.5 \\ 
\hline
\textbf{YOLO4}\cite{dang2023yoloweeds} & -- & -- & 94.78 & 95.04 & 95.22 & 89.48 & $\sim$66M & $\sim$141 \\

\hline
\end{tabular}%
}
\label{tab:EcoWeedNet_performance_SimAM_SPAB}
\end{table*}

\begin{figure*}[t]
\centering
\setlength{\fboxsep}{0pt} 
\setlength{\fboxrule}{0.5pt} 

\begin{minipage}[c]{0.02\textwidth}
    \rotatebox{90}{Original}
\end{minipage}%
\begin{minipage}[c]{\textwidth}
    \fbox{\includegraphics[width=0.12\textwidth]{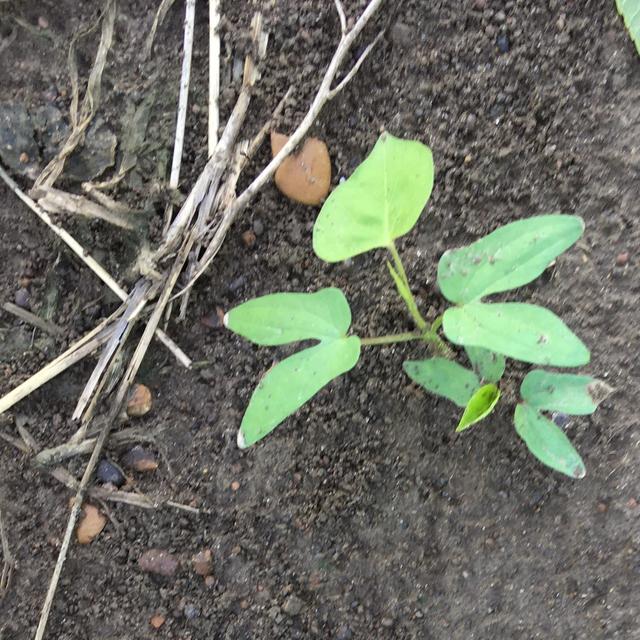}}%
    \fbox{\includegraphics[width=0.12\textwidth]{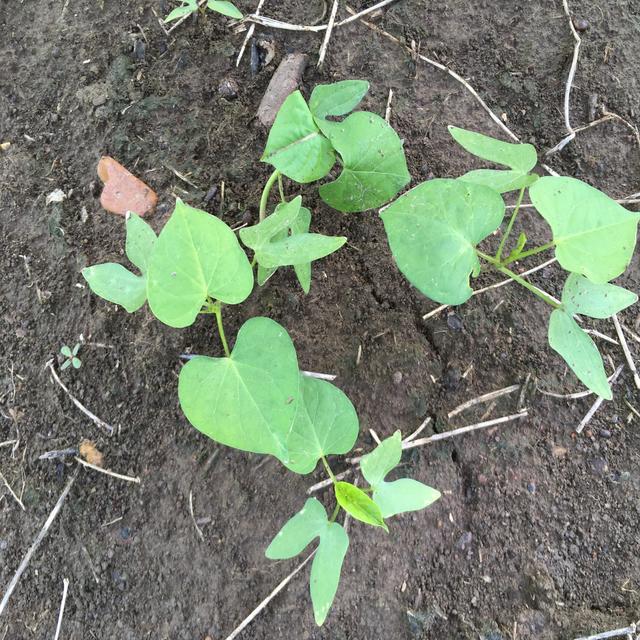}}%
    \fbox{\includegraphics[width=0.12\textwidth]{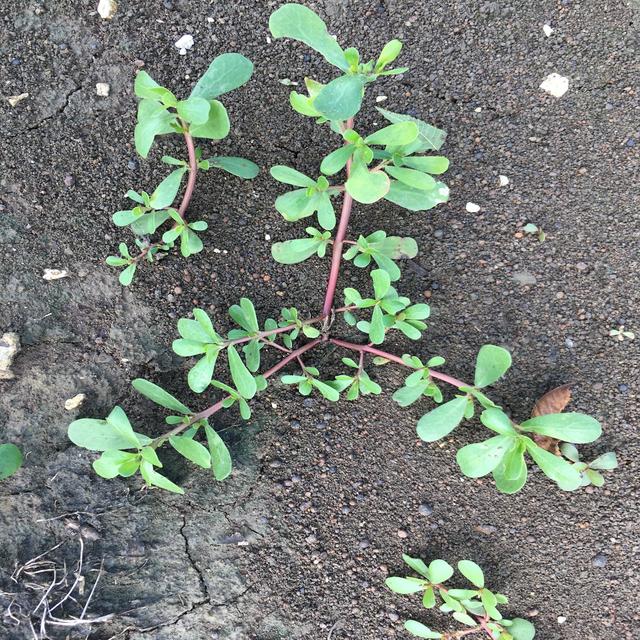}}%
    \fbox{\includegraphics[width=0.12\textwidth]{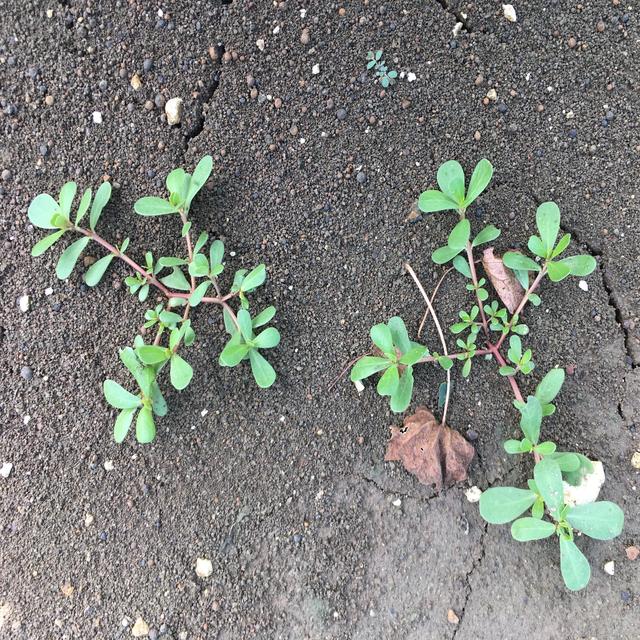}}%
    \fbox{\includegraphics[width=0.12\textwidth]{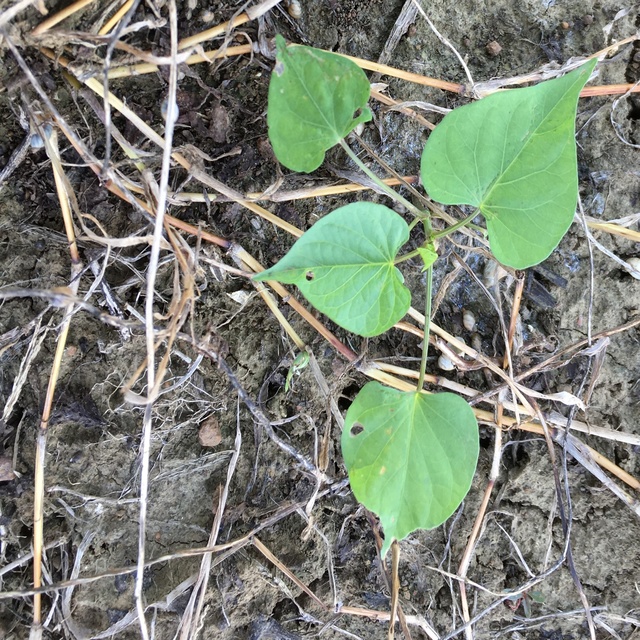}}%
    \fbox{\includegraphics[width=0.12\textwidth]{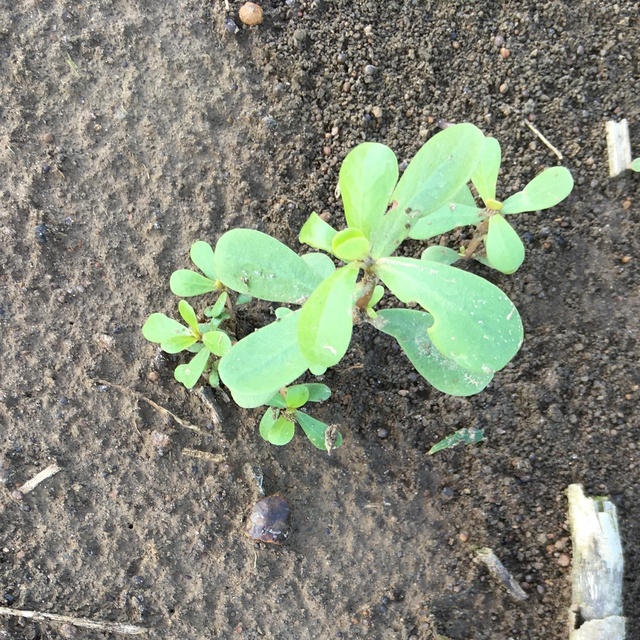}}%
    \fbox{\includegraphics[width=0.12\textwidth]{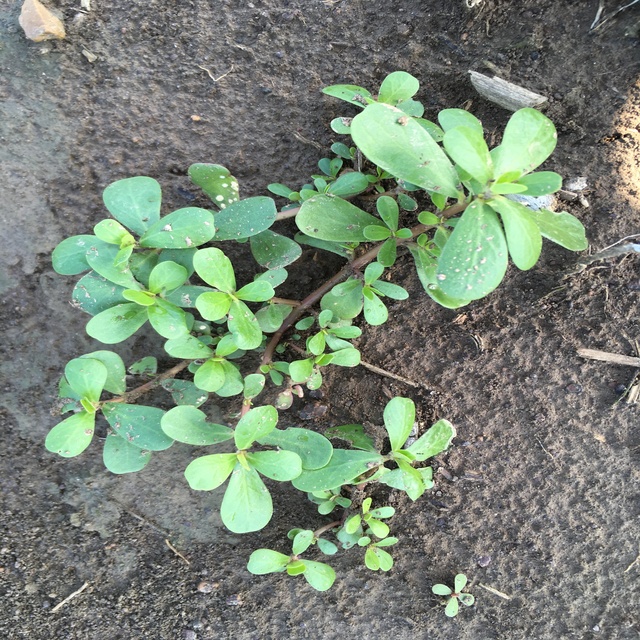}}%
    \fbox{\includegraphics[width=0.12\textwidth]{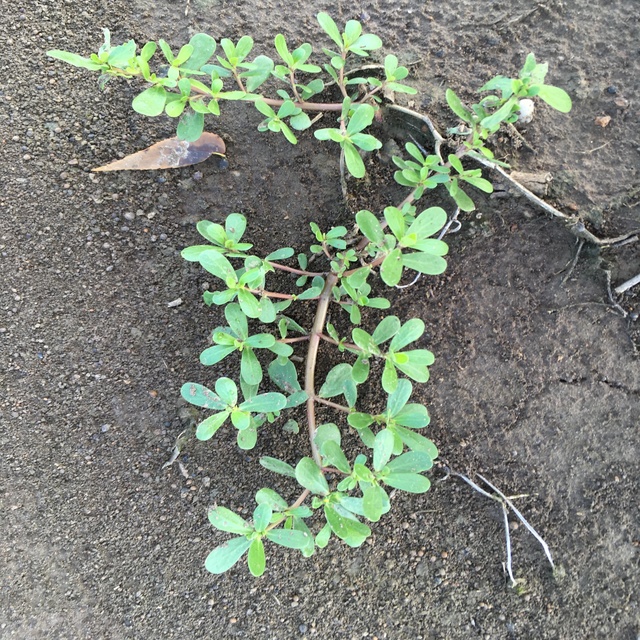}}
\end{minipage}

\vspace{0.07 cm} 

\begin{minipage}[c]{0.02\textwidth}
    \rotatebox{90}{YOLOv11n}
\end{minipage}%
\begin{minipage}[c]{\textwidth}
    \fbox{\includegraphics[width=0.12\textwidth]{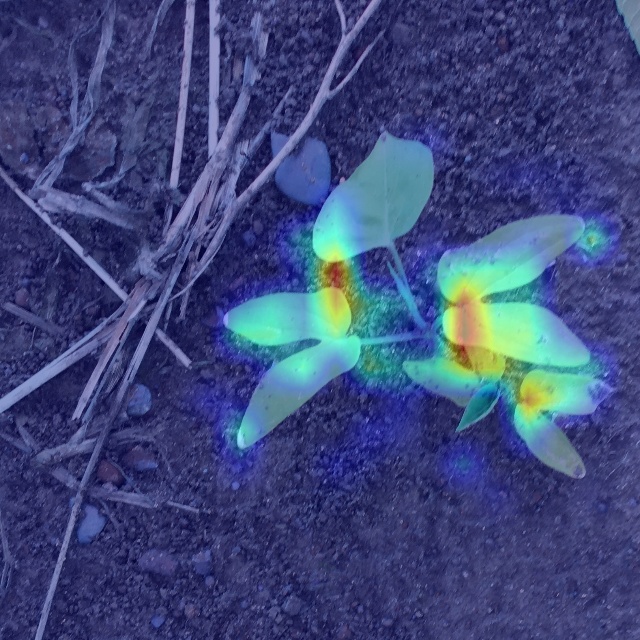}}%
    \fbox{\includegraphics[width=0.12\textwidth]{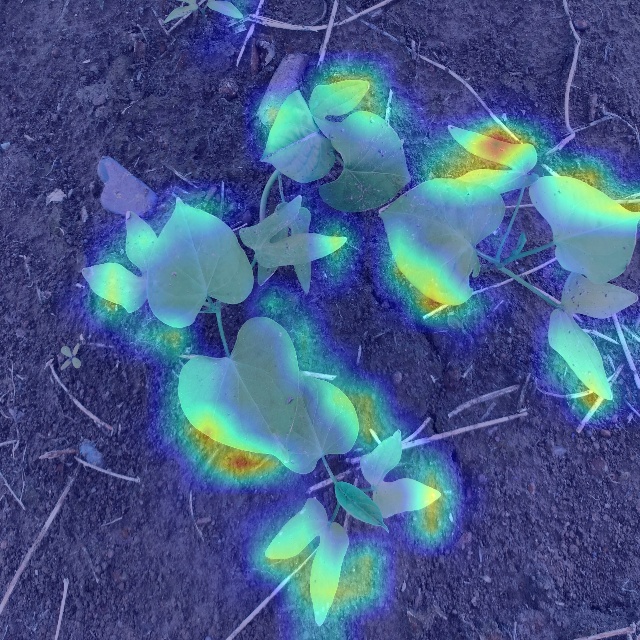}}%
    \fbox{\includegraphics[width=0.12\textwidth]{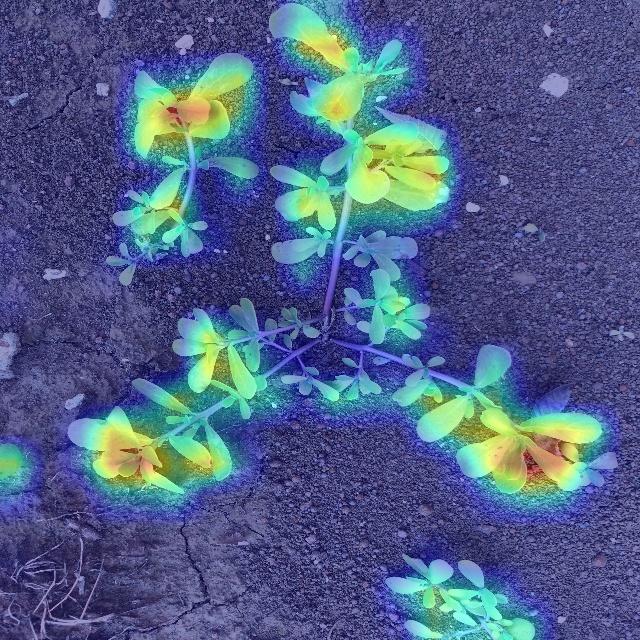}}%
    \fbox{\includegraphics[width=0.12\textwidth]{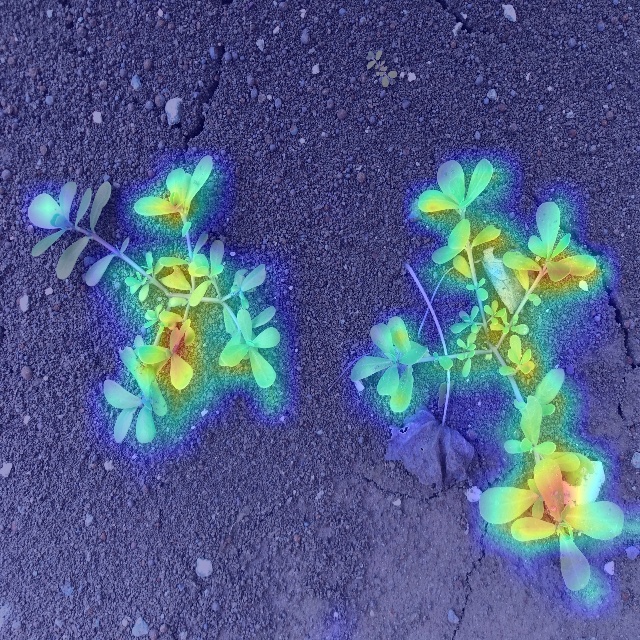}}%
    \fbox{\includegraphics[width=0.12\textwidth]{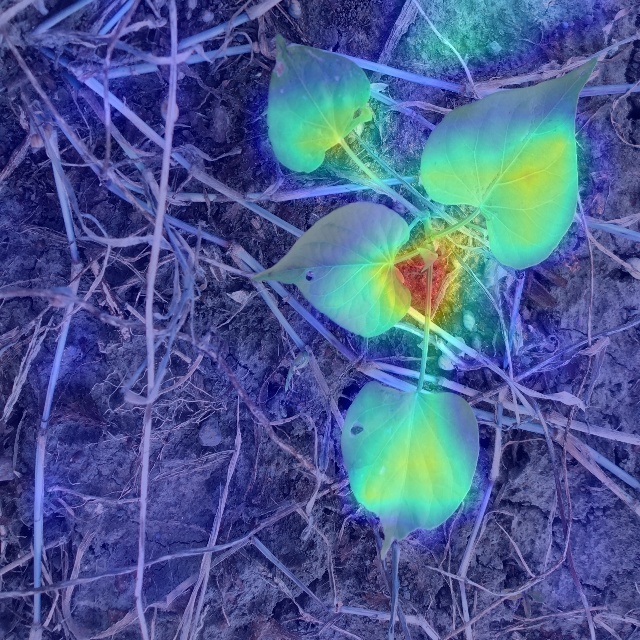}}%
    \fbox{\includegraphics[width=0.12\textwidth]{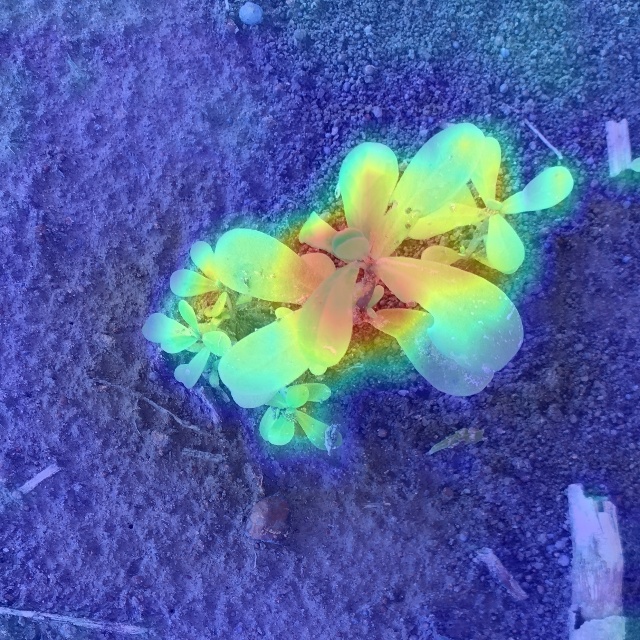}}%
    \fbox{\includegraphics[width=0.12\textwidth]{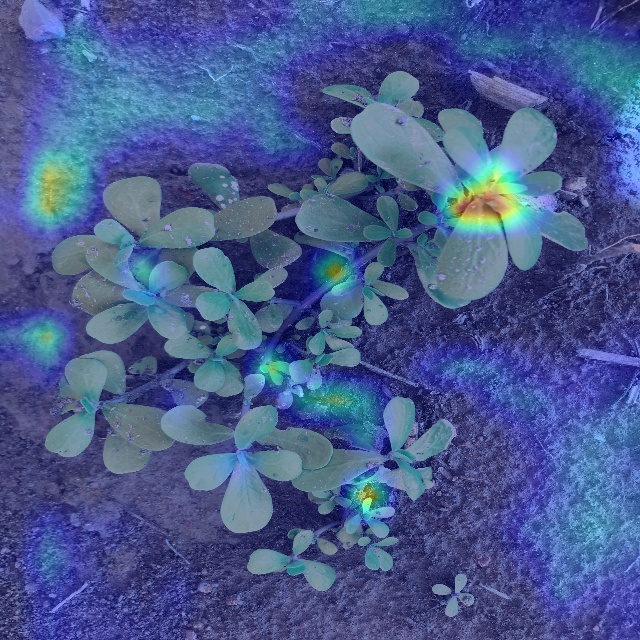}}%
    \fbox{\includegraphics[width=0.12\textwidth]{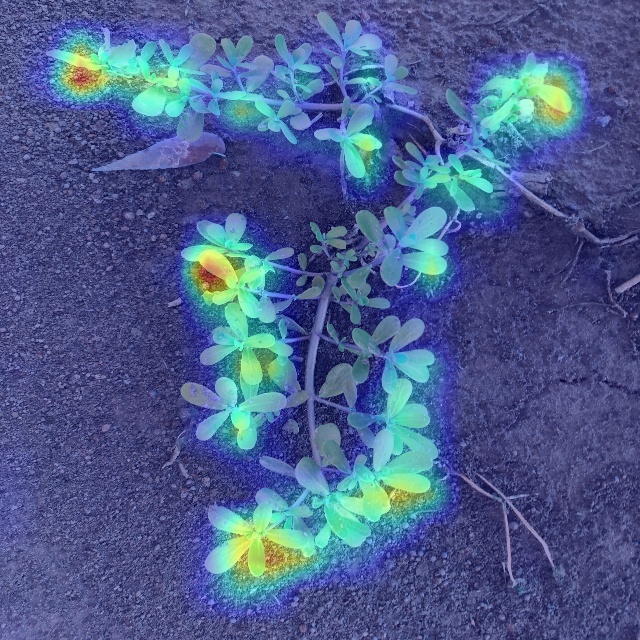}}
\end{minipage}

\vspace{0.07 cm} 

\begin{minipage}[c]{0.02\textwidth}
    \rotatebox{90}{YOLOv12n}
\end{minipage}%
\begin{minipage}[c]{\textwidth}
    \fbox{\includegraphics[width=0.12\textwidth]{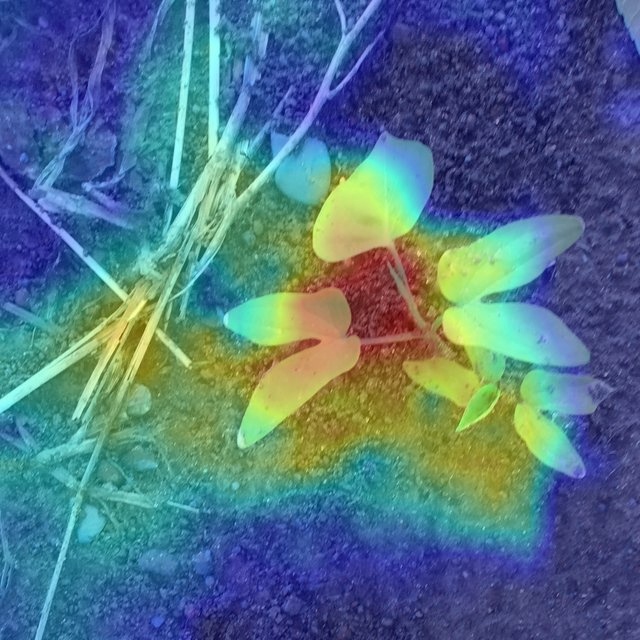}}%
    \fbox{\includegraphics[width=0.12\textwidth]{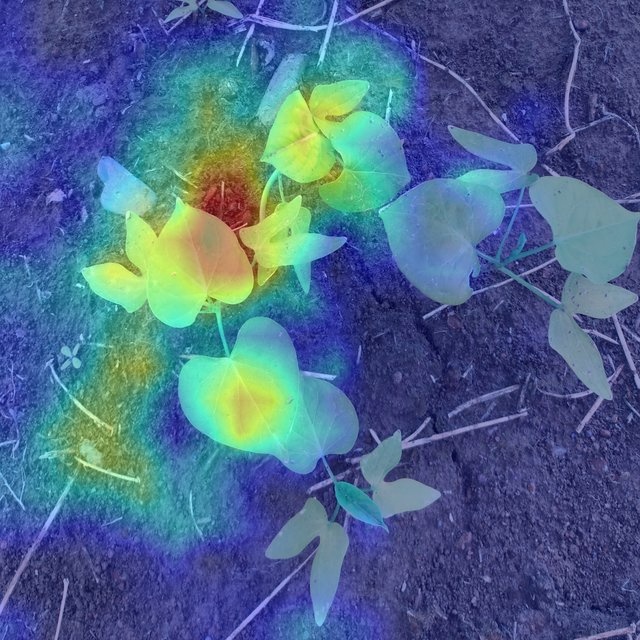}}%
    \fbox{\includegraphics[width=0.12\textwidth]{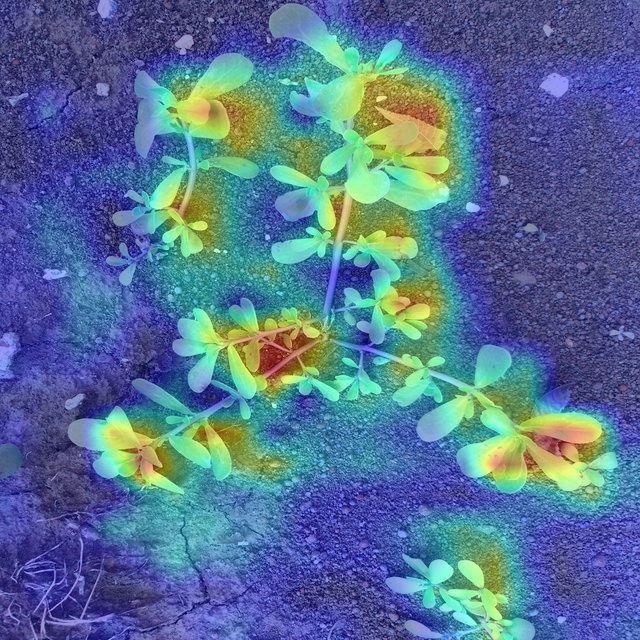}}%
    \fbox{\includegraphics[width=0.12\textwidth]{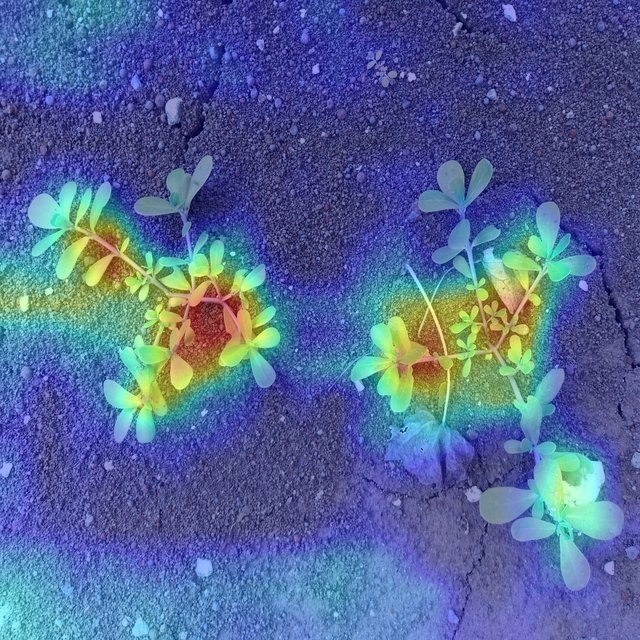}}%
    \fbox{\includegraphics[width=0.12\textwidth]{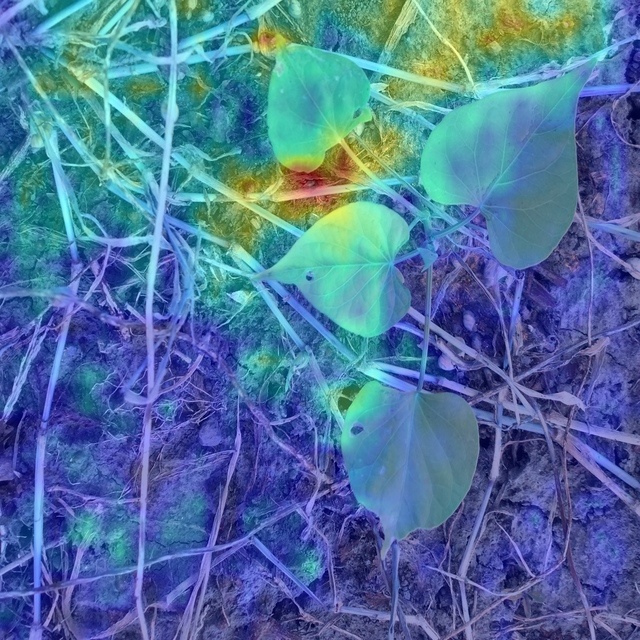}}%
    \fbox{\includegraphics[width=0.12\textwidth]{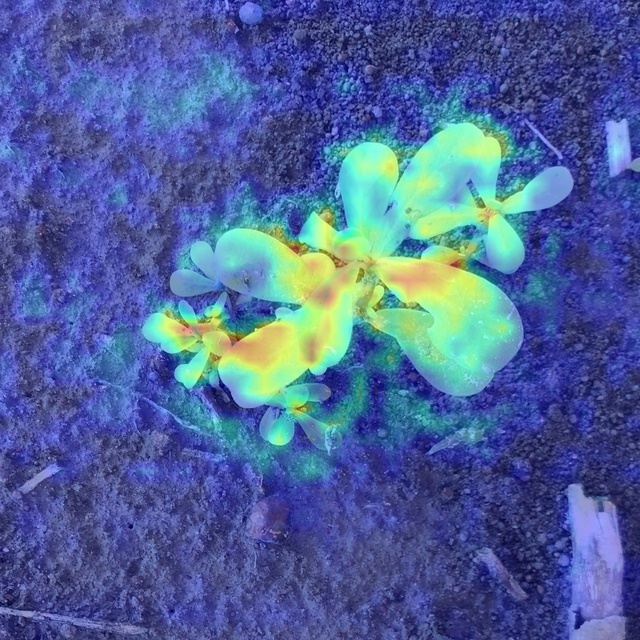}}%
    \fbox{\includegraphics[width=0.12\textwidth]{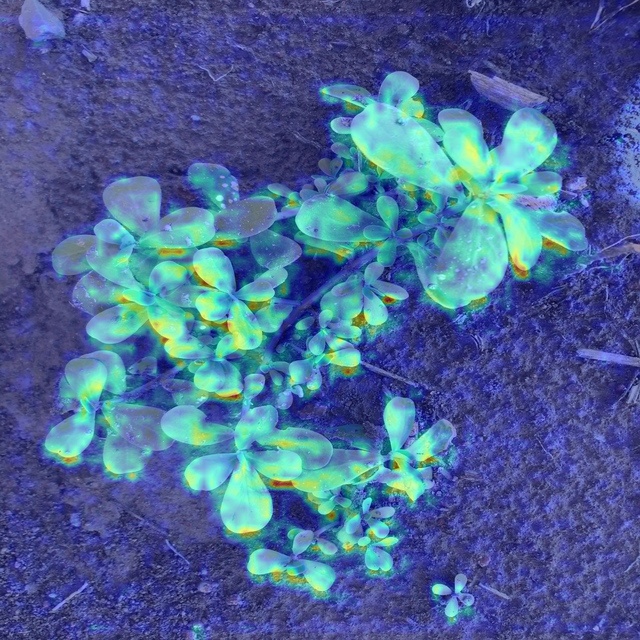}}%
    \fbox{\includegraphics[width=0.12\textwidth]{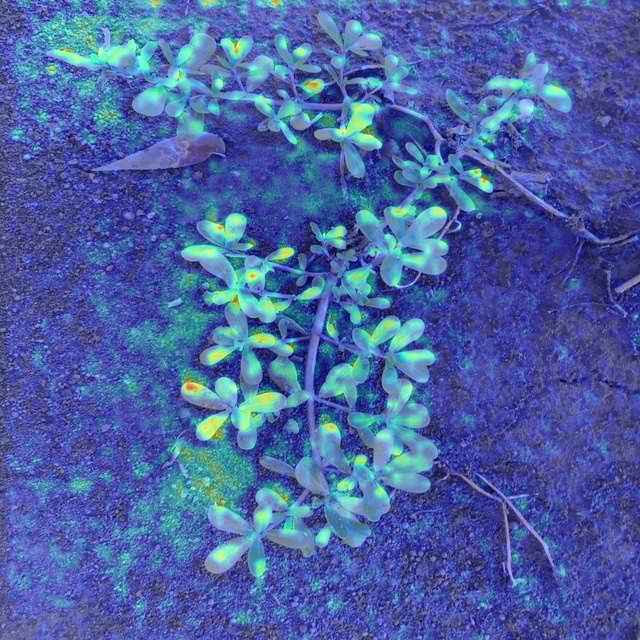}}
\end{minipage}

\vspace{0.07 cm} 

\begin{minipage}[c]{0.02\textwidth}
    \rotatebox{90}{EcoWeedNet}
\end{minipage}%
\begin{minipage}[c]{\textwidth}
    \fbox{\includegraphics[width=0.12\textwidth]{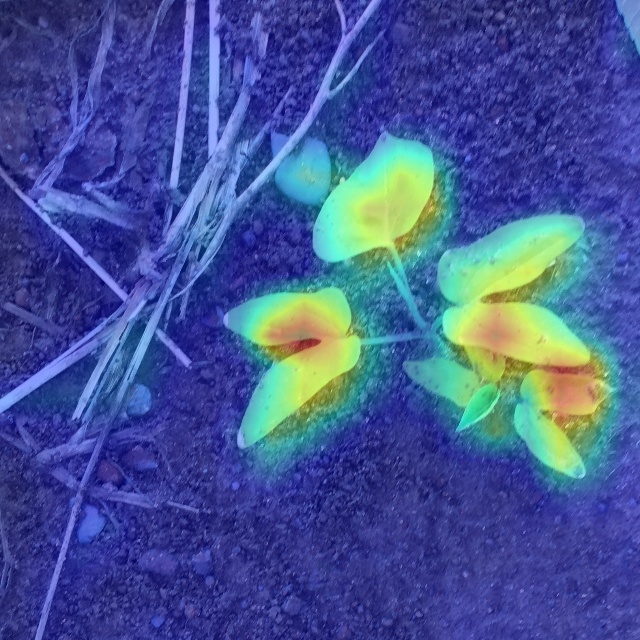}}%
    \fbox{\includegraphics[width=0.12\textwidth]{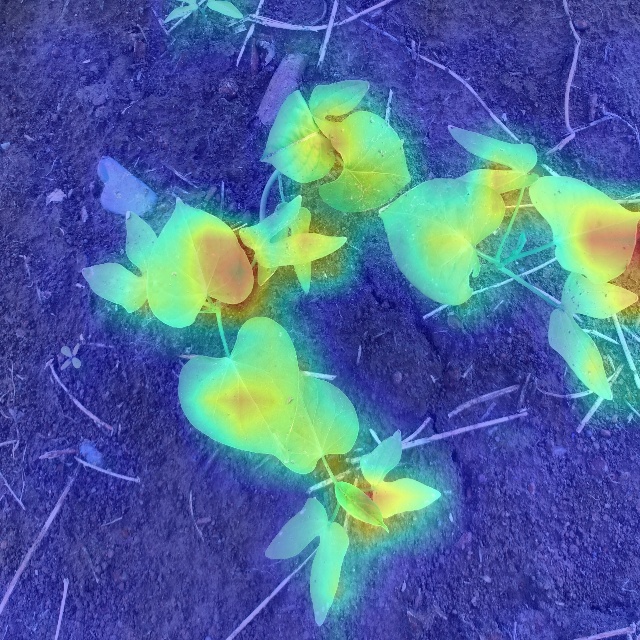}}%
    \fbox{\includegraphics[width=0.12\textwidth]{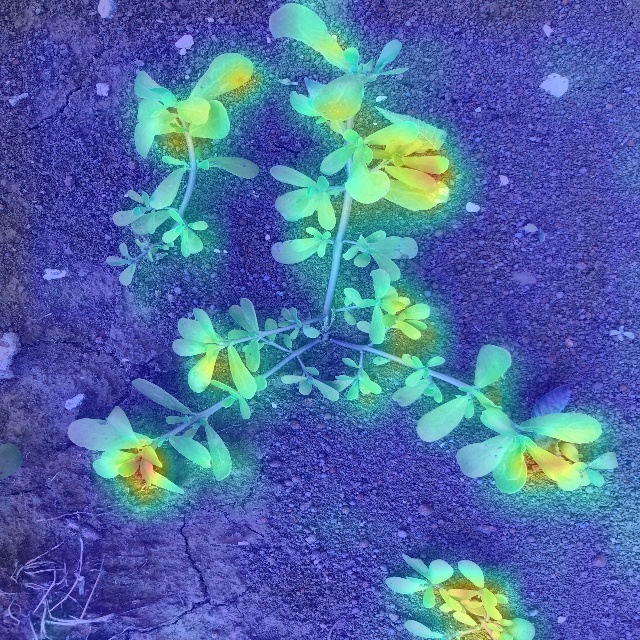}}%
    \fbox{\includegraphics[width=0.12\textwidth]{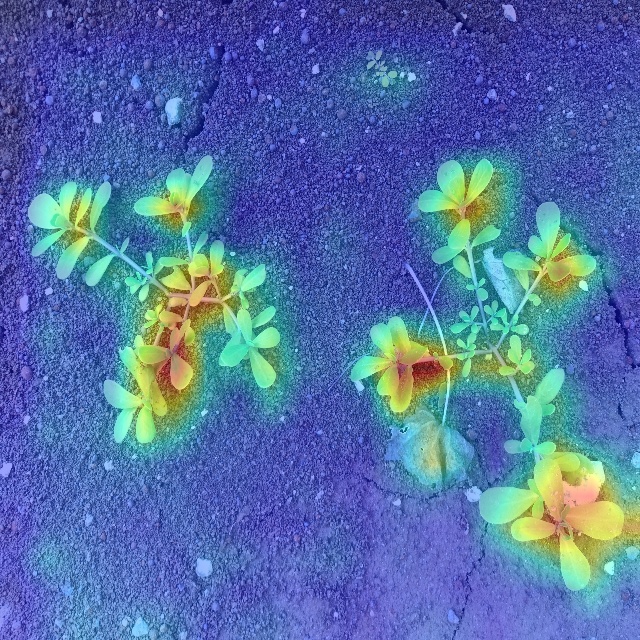}}%
    \fbox{\includegraphics[width=0.12\textwidth]{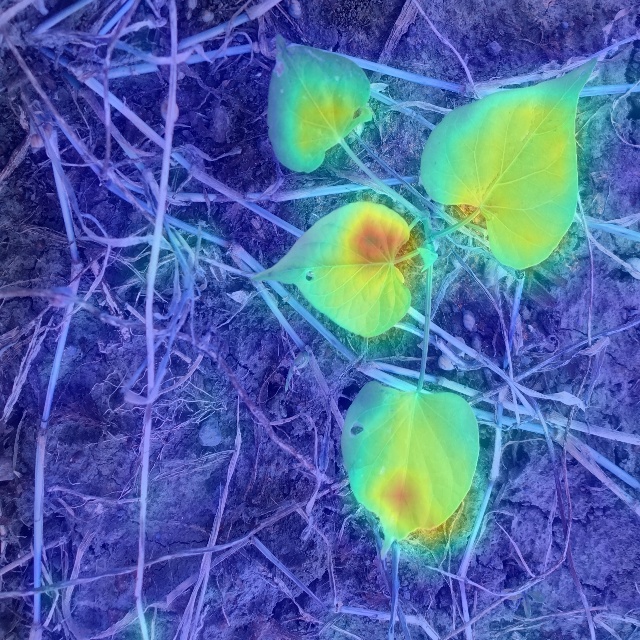}}%
    \fbox{\includegraphics[width=0.12\textwidth]{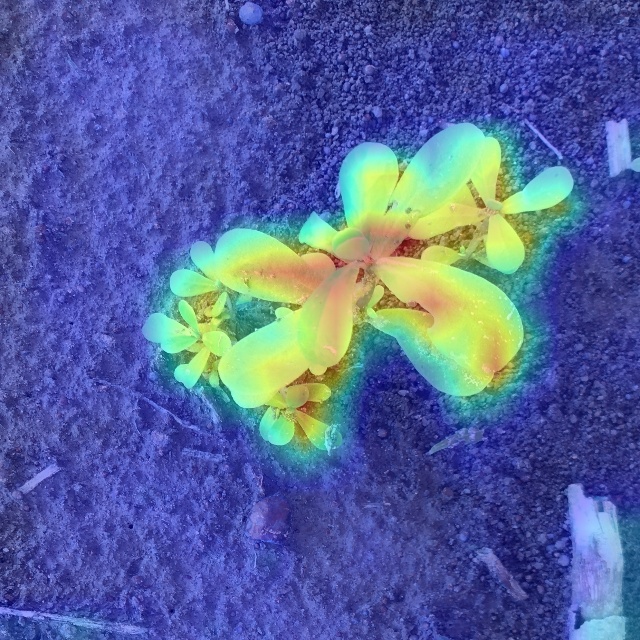}}%
    \fbox{\includegraphics[width=0.12\textwidth]{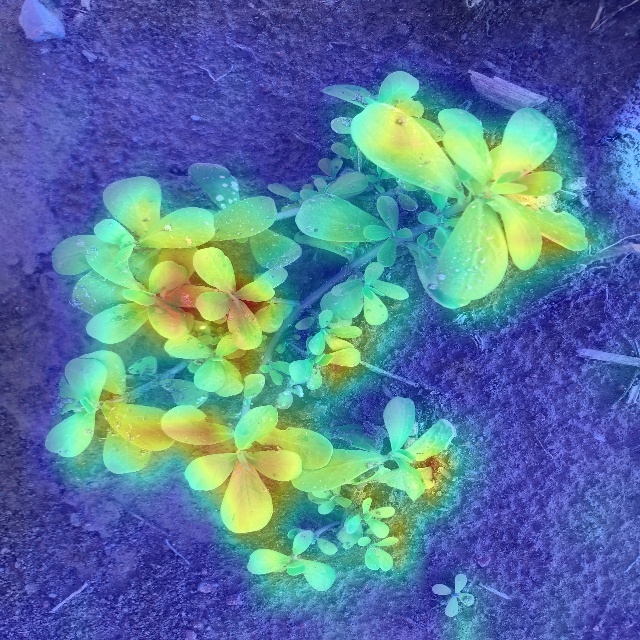}}%
    \fbox{\includegraphics[width=0.12\textwidth]{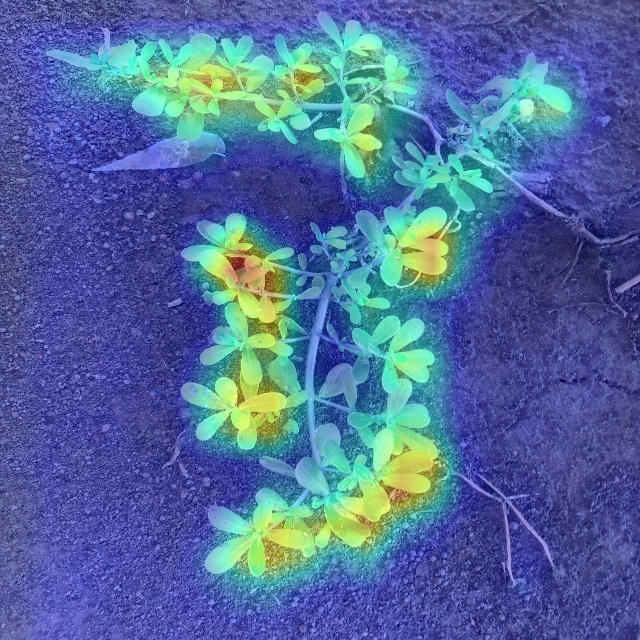}}
\end{minipage}

\caption{Comparative GradCAM$++$ visualizations for the baseline YOLOv11n and EcoWeedNet feature responses. The proposed EcoWeedNet demonstrated better emphasis and focus on weed-relevant regions.}

\label{fig:image_comparison}
\end{figure*}

\section{Results}

\begin{figure*}[t]
    \centering

    \begin{minipage}{0.4\textwidth}
        \includegraphics[width=\linewidth]{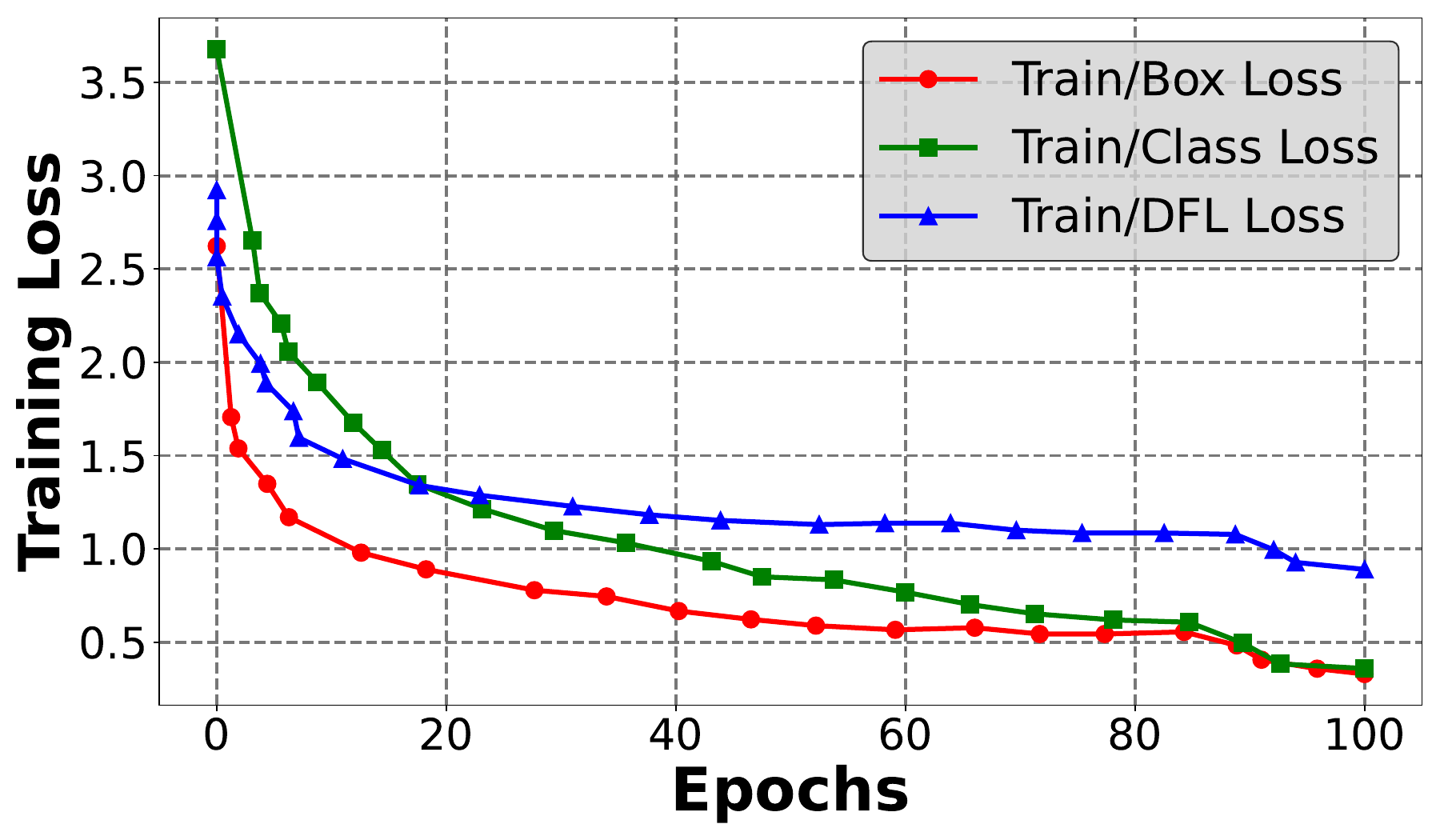}
        \subcaption{Training losses}
        \label{fig:train_losses}
    \end{minipage}\hfill
    \begin{minipage}{0.4\textwidth}
        \includegraphics[width=\linewidth]{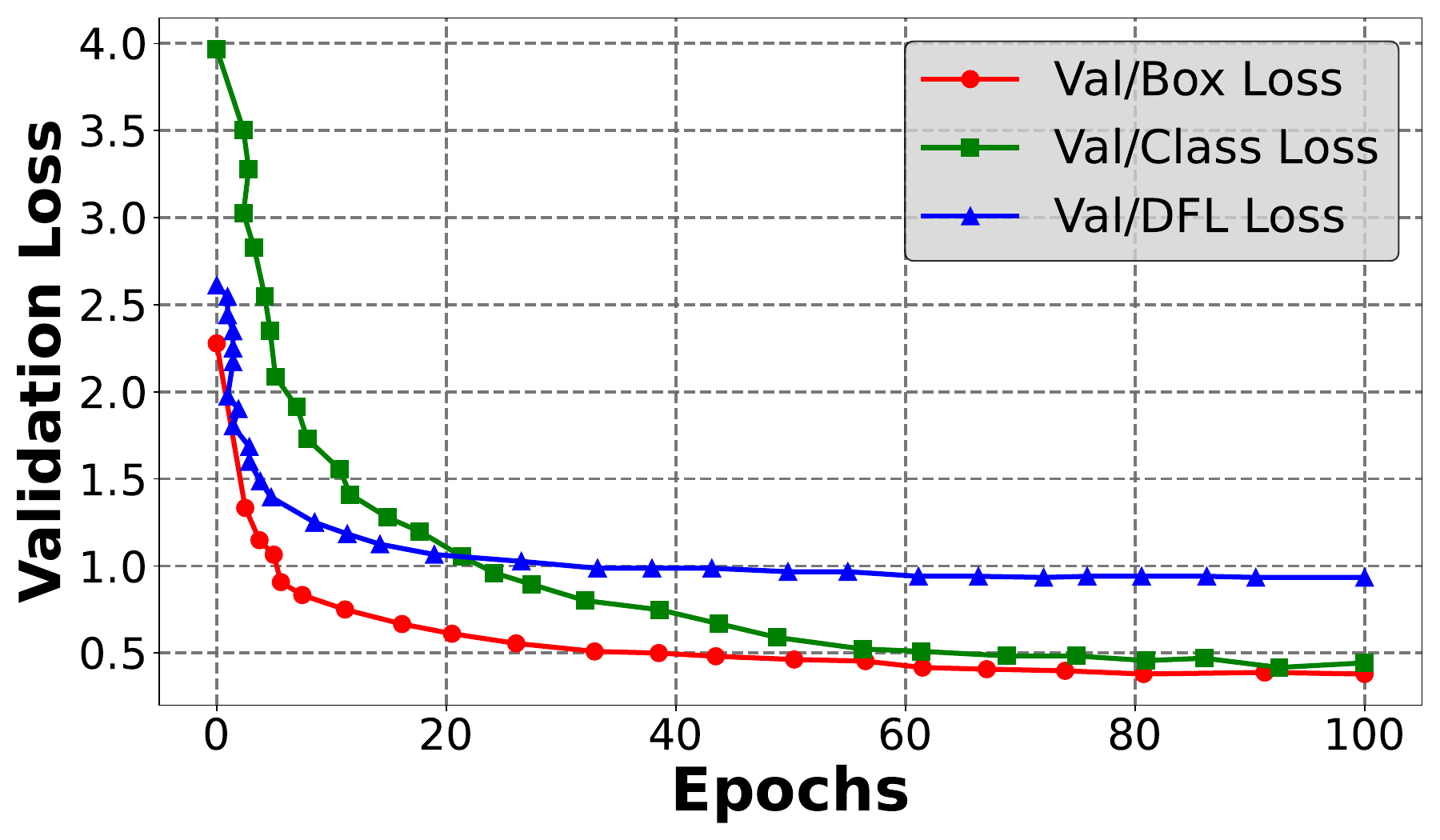}
        \subcaption{Validation losses}
        \label{fig:val_losses}
    \end{minipage}

    \vspace{0.3cm}  

    \begin{minipage}{0.4\textwidth}
        \includegraphics[width=\linewidth]{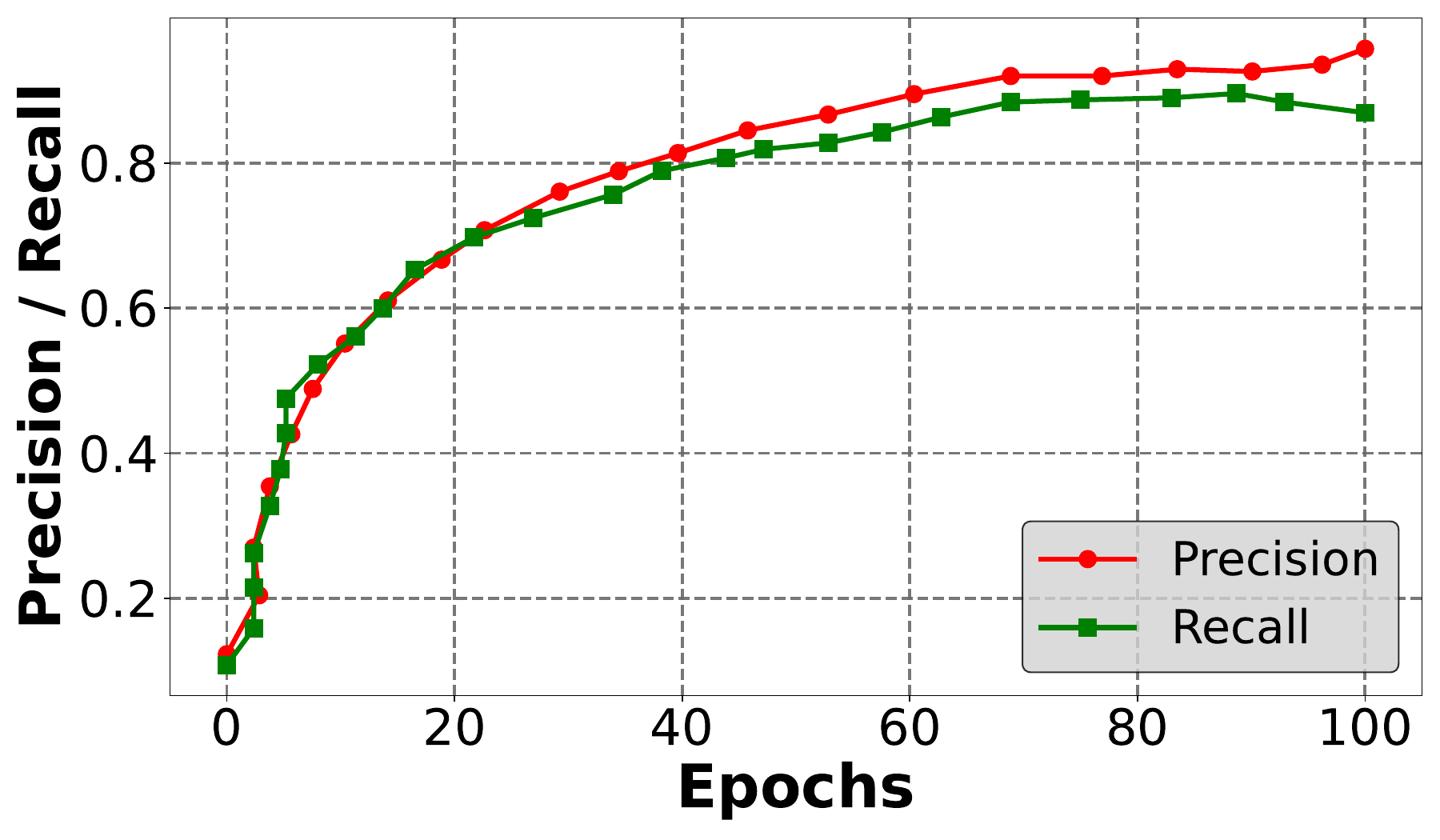}
        \subcaption{Precision and Recall}
        \label{fig:train_PR}
    \end{minipage}\hfill
    \begin{minipage}{0.4\textwidth}
        \includegraphics[width=\linewidth]{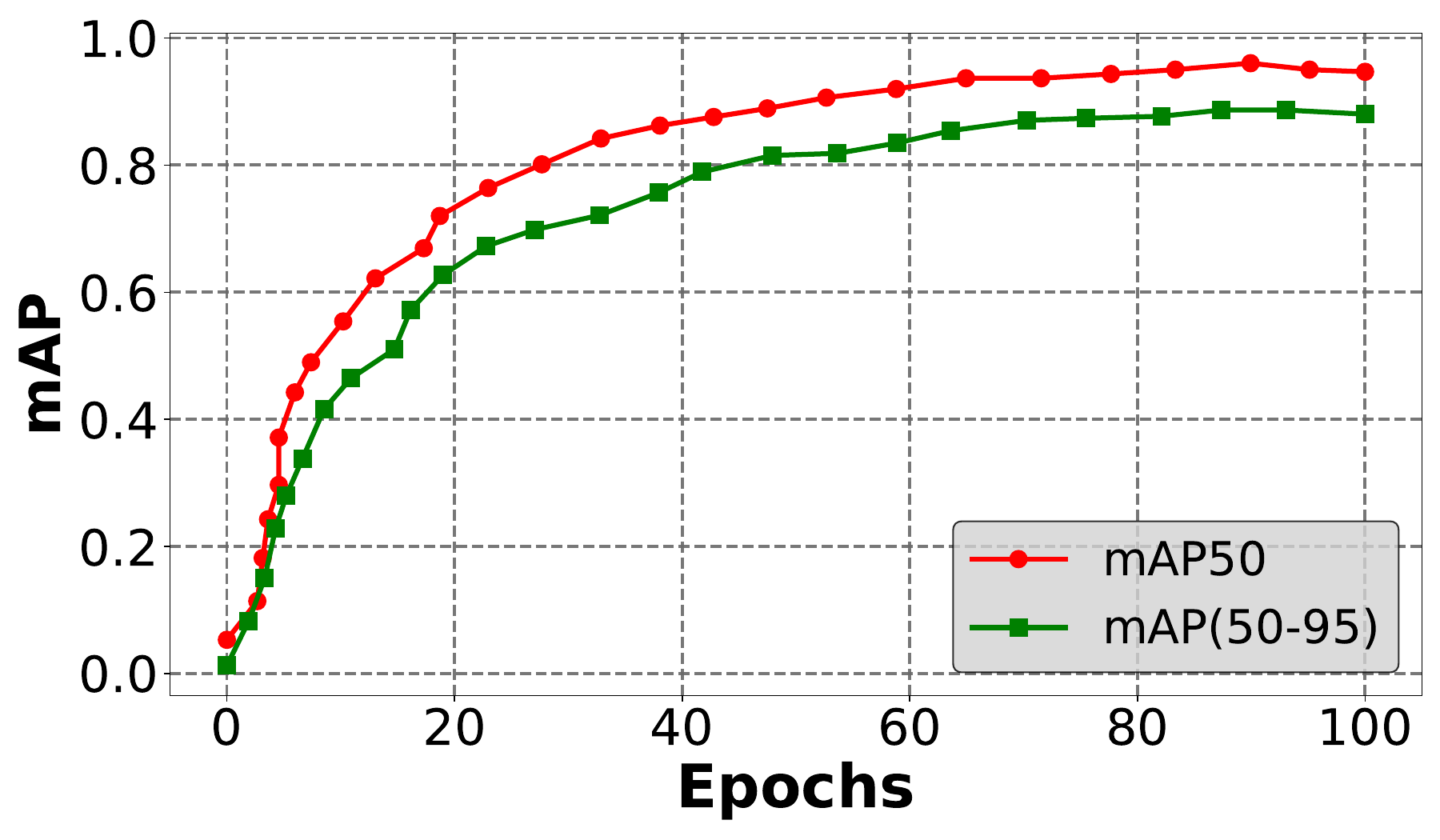}
        \subcaption{Mean Average Precision}
        \label{fig:train_mAP}
    \end{minipage}

    \caption{EcoWeedNet's performance metrics over 100 epochs. 
    \textbf{Box Loss} refers to the localization error in bounding box regression, 
    \textbf{Class Loss} measures classification errors, and 
    \textbf{DFL Loss} (Distribution Focal Loss) enhances bounding box precision. 
    \textbf{Precision} and \textbf{Recall} evaluates classification performance, 
    while \textbf{mAP50} and \textbf{mAP(50–95)} indicate detection accuracy at different IoU thresholds.}
    \label{fig:comprehensive_metrics}
\end{figure*}

\vspace{-0.2cm}

\subsection{Performance of the proposed model}

The enhanced nano model contributes significantly to the field of precision agriculture by providing a reliable, efficient, and economically feasible solution for weed detection. Our proposed model combines the capabilities of big architectures with low computational cost, as shown in Figure~\ref{fig:comprehensive_metrics}, highlighting the importance and suitability of our model for real-time consumer electronic applications. Investigating EcoWeedNet architecture and integrating SPAB and SimAM modules in the backbone and the neck enhances feature discrimination without adding substantial complexity while maintaining a lightweight network.

The model size (parameters) and computational cost (GFLOPs) are taken as metrics to quantify EcoWeedNet's lightweight characteristic. For clarity, they are included alongside detection performance in Table \ref{tab:EcoWeedNet_performance_SimAM_SPAB} specifically to emphasize the model's efficiency and deployment readiness in low-resource devices.

The introduced enhancements add only a tiny overhead compared to the substantial improvement achieved in detection performance. Per-class results are shown in Table~\ref{tab:per_class_metrics}. Class imbalance was addressed using Distributed Focal Loss (DFL), enabling learning to focus on minority weed classes.

The mAP$_{50-95}$ metric evaluates the performance of a model at different IoU thresholds (0.50 to 0.95), a more conservative metric compared to mAP$_{50}$, which evaluates the performance at a single IoU threshold of 0.50. Thus, the mAP$_{50-95}$ is usually lower due to the greater difficulty in accurately predicting bounding boxes at higher IoUs.

\begin{table}[H]
\centering
\caption{Per-class Precision, Recall, and F1-score of EcoWeedNet}
\label{tab:per_class_metrics}
\resizebox{\columnwidth}{!}{
\begin{tabular}{cccc}
\hline
\textbf{Class} & \textbf{Precision (P)} & \textbf{Recall (R)} & \textbf{F1-score} \\ \hline
Eclipta & 0.898 & 0.844 & 0.870 \\ 
Ipomoea indica & 0.969 & 0.944 & 0.956 \\ 
Eleusine indica & 0.987 & 0.956 & 0.971 \\ 
Sida rhombifolia & 0.967 & 0.888 & 0.926 \\ 
Physalis angulata & 0.975 & 0.920 & 0.947 \\ 
Senna obtusifolia & 0.941 & 0.877 & 0.908 \\ 
Amaranthus palmeri & 0.975 & 0.870 & 0.920 \\ 
Euphorbia maculata & 0.955 & 0.867 & 0.909 \\ 
Portulaca oleracea & 0.934 & 0.900 & 0.917 \\ 
Mollugo verticillata & 0.940 & 0.913 & 0.926 \\ 
Amaranthus tuberculatus & 0.965 & 0.965 & 0.965 \\ 
Ambrosia artemisiifolia & 0.957 & 0.871 & 0.912 \\ \hline
\textbf{Average} & \textbf{0.952} & \textbf{0.889} & \textbf{0.919} \\ \hline
\end{tabular}}
\end{table}

\subsection{Performance of EcoWeedNet excluding SPAB}

In our study, we investigated the performance of EcoWeedNet after excluding the SPAB module to highlight the impact of its absence on detection capabilities. Table~\ref{tab: EcoWeedNet_performance_SimAM} shows the resulting metrics.

\begin{table}[H]
\centering
\caption{Performance of EcoWeedNet Excluding SPAB}
\resizebox{\columnwidth}{!}{%
\begin{tabular}{ccccccc}
\hline
\textbf{SimAM Index} & \textbf{Precision (\%)} & \textbf{Recall (\%)} & \textbf{mAP50 (\%)} & \textbf{mAP(50-95) (\%)} & \textbf{Param.} & \textbf{GFLOPs} \\ 
\hline
1 & 89.9   & 88.8  & 94.3  & 87.5  & 2.6M  & 6.5  \\ 

1,3  & 89.5   & 87.4  & 92.9  & 86.7  & 2.6M  & 6.5  \\ 

1,9 & 90.734 & \textbf{90.5} & 94 & \textbf{88.2} & 2.6M & 6.5 \\ 

1,9,13  & 92.9   & 86.1  & 94.1  & 88.1  & 2.6M  & 6.5  \\ 

1,13,17 & 92.3   & 85.7  & 93.6  & 87  & 2.6M  & 6.5  \\

11,15   & 93.3   & 89.5  & \textbf{94.5}  & 88.1  & 2.6M  & 6.5  \\

1,3,13,17 & \textbf{94.3} & 87.5  & 94.1  & 87.8  & 2.6M  & 6.5  \\

1,3,14,18 & 92.0   & 87.7  & 93.7  & 87.2  & 2.6M  & 6.5  \\
\hline
\end{tabular}%
}
\label{tab: EcoWeedNet_performance_SimAM}
\end{table}

\subsection{Performance of EcoWeedNet excluding SimAM}

The absence of the SimAM module negatively impacts the model's performance, leading to reduced precision and mAP, as shown in Table~\ref{tab: EcoWeedNet_performance_SPAB}. Emphasizing its essential role in refining feature representation and enhancing detection accuracy.

\begin{table}[H]
\centering
\caption{Performance of EcoWeedNet Excluding SimAM}
\resizebox{\columnwidth}{!}{%
\begin{tabular}{ccccccc}
\hline
\textbf{SPAB Index} & \textbf{Precision (\%)} & \textbf{Recall (\%)} & \textbf{mAP50 (\%)} & \textbf{mAP(50-95) (\%)} & \textbf{Parameters} & \textbf{GFLOPs} \\ 
\hline
1 & \textbf{92.9} & 87.5 & \textbf{94.4} & \textbf{88.4} & \textbf{2.63M} & \textbf{7.9} \\ 
3 & 89.8 & \textbf{89.8} & 94.0  & 87.3 & 2.74M & \textbf{7.9}  \\ 
5 & 90.0 & 88.6 & 93.8 & 87.6 & 5.05M & 12.1  \\ 
\hline
\end{tabular}%
}
\label{tab: EcoWeedNet_performance_SPAB}
\end{table}

\subsection{GradCAM-based Explainability Analysis}

The proposed EcoWeedNet provides much better performance and the model's ability to detect the most relevant areas as shown by GradCAM++ heatmaps, as demonstrated in Figure~\ref{fig:image_comparison}. These examples indicate EcoWeedNet's better feature emphasis and focus on relevant regions compared to the baseline YOLOv11n; in this way, interpretability improves along with the model's capacity to analyze complex surroundings.

\vspace{-0.2cm}

\subsection{K-Fold Cross-Validation}

We've performed 5-fold cross-validation on 90\% of data (validation set + training set), with the rest of data (10\%) set aside for separate use in the testing. This puts the model's performance under a more precise test for generalized performance while reducing the risk of biased results.

We present the performance metrics averaged across all the folds after performing the experiments in a 5-fold cross-validation manner. The test performance was obtained when the EcoWeedNet model was trained using the training+validation data set of 90\% and tested on the held-out test set of 10\%.
The cross-validation results are summarized in Table~\ref{tab:kfold_results}.

\begin{table}[H]
\centering
\caption{5-Fold Cross-Validation Results of EcoWeedNet}
\label{tab:kfold_results}
\resizebox{\columnwidth}{!}{%
\begin{tabular}{ccccc}
\hline
\textbf{Fold} & \textbf{Precision (\%)} & \textbf{Recall (\%)} & \textbf{mAP@50 (\%)} & \textbf{mAP(50–95) (\%)} \\ \hline
Fold 1   &  91.4 & 88.7 & 94.9 & 88.8  \\ 
Fold 2   &  90.8 & 89.1 & 94.4 & 88.5   \\
Fold 3   &  90.4 & 88.3 & 95.1 & 89.0   \\
Fold 4   &  89.8 & 87.9 & 94.3 & 87.8   \\
Fold 5   &  91.1 & 87.4 & 93.4 & 88.7   \\ \hline
\textbf{Average} & \textbf{90.7} & \textbf{88.3} & \textbf{94.4} & \textbf{88.6}   \\ \hline
\textbf{Std Dev} & \textbf{0.58} & \textbf{0.65} & \textbf{0.61} & \textbf{0.42}   \\ \hline
\end{tabular}%
}
\end{table}

\subsection{Impact of Training Set Size on Performance and Training Time}

We performed three experiments to identify how training set size affects the efficiency and performance of models. In Experiment 1, we trained using 100\% of the overall training dataset, in the second experiment, we trained using 50\% of the training dataset, and in the third experiment trained using 25\% of the training dataset. While the three experiments were done on the same 10\% of the testing dataset. This enabled us to trade in performance degradation against savings in the training dataset size.

We implemented the three experiments using EcoWeedNet, using the same training hyperparameters in all experiments. The aim was to analyze the impact of reducing the training dataset size on detection performance.

\begin{table}[H]
\centering
\caption{Comparison of EcoWeedNet performance using different training set sizes}
\label{tab:dataset_size_impact}
\resizebox{\columnwidth}{!}{%
\begin{tabular}{ccccc}
\hline
\textbf{Experiments} & \textbf{Precision (\%)} & \textbf{Recall (\%)} & \textbf{mAP@50 (\%)} & \textbf{mAP(50–95) (\%)} \\ \hline
Exp. 1  & 93.2 & 89   & 95.2 & 88.9  \\ 
Exp. 2  & 85.4 & 79.5 &	86.1 & 78.3  \\
Exp. 3  & 80   & 68.5 &	77   & 67.2  \\ \hline
\end{tabular}%
}
\end{table}

These results highlight that reducing the training data leads to a noticeable drop in performance. Therefore, the size of the training dataset is crucial to maximizing the evaluation metrics from the EcoWeedNet model.

\subsection{Comparative Analysis}

In direct comparison with the well-established YOLOv4 \cite{dang2023yoloweeds}, our enhanced nano model achieved a competitive performance, scoring a mAP50 of up to 95.2\% and mAP(50-95) of 88.9\% on the same dataset. These results outperformed the YOLO12n and are remarkably close to those of YOLOv4, which scores a mAP50 of 95.22\% and mAP(50-95) of 89.48\%, as shown in Table~\ref{tab:EcoWeedNet_performance_SimAM_SPAB}. However, YOLOv4 consumes significantly higher computational resources, approximately 66M parameters and 141 GFLOPs, compared to our model's maximum of 2.78M parameters and 9.3 GFLOPs.

\begin{figure}[H]
    \centering
    \setlength{\fboxsep}{0pt} 
    \setlength{\fboxrule}{1pt} 
    \fbox{\includegraphics[width=\columnwidth]{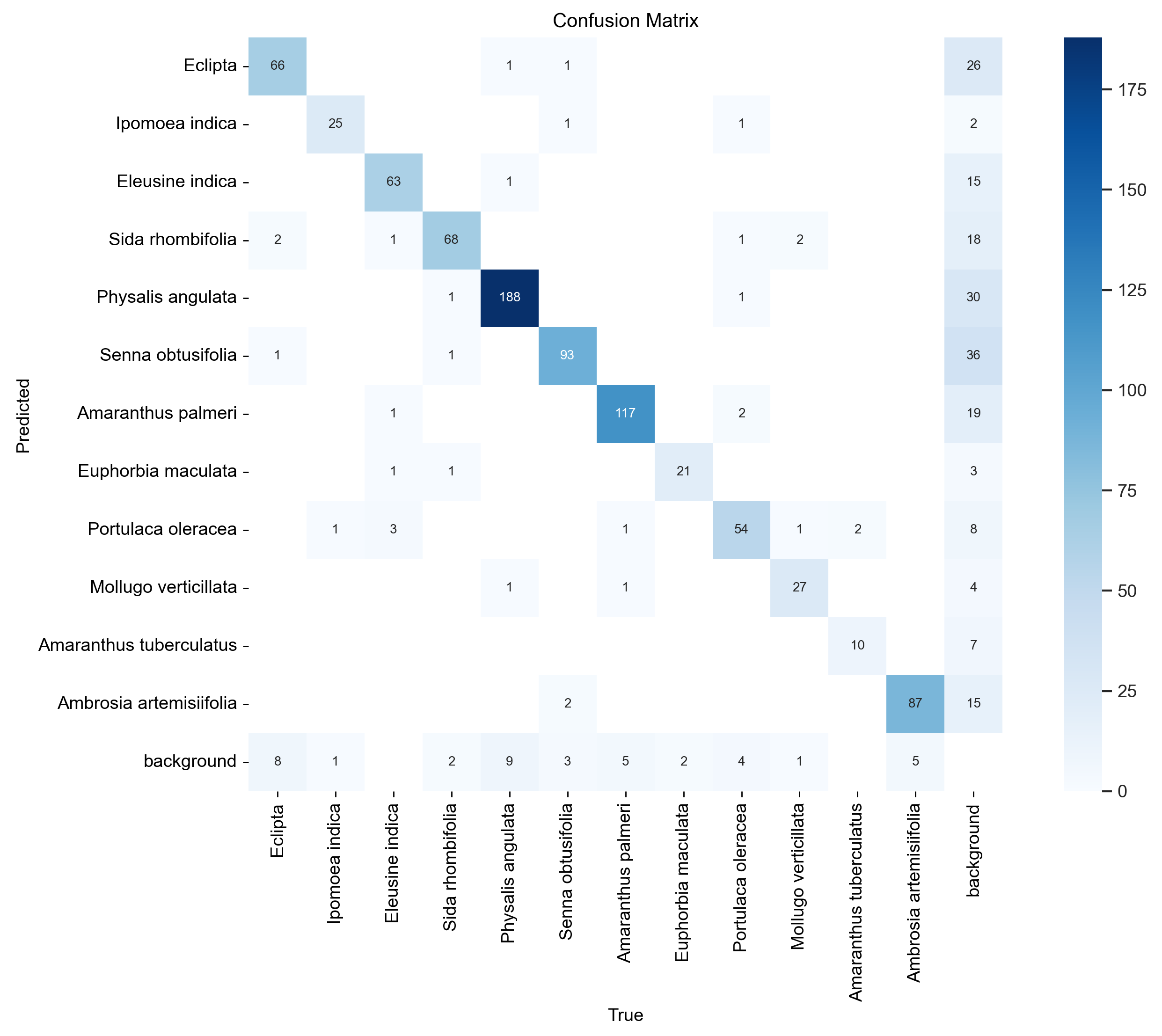}}
    \caption{Confusion matrix of the EcoWeedNet model demonstrating the number of correct multi-class classifications and low inter-class confusion.}
    \label{fig:our_confusion_matrix}
\end{figure}

While the YOLOv4 showed a slightly higher mAP score, it is unsuitable for consumer electronic devices. The proposed EcoWeedNet has demonstrated exceptional performance and efficiency; hence, it is ideal for practical agricultural applications with minimum energy consumption. Figure~\ref{fig:our_confusion_matrix} illustrates the confusion matrix, emphasizing our model's capability to accurately distinguish among the different weed classes and visualize its classification performance clearly.

\section{Conclusion}

This work proposed the EcoWeedNet model to enhance weed detection capabilities without introducing significant computational complexity for next-generation sustainable agricultural consumer electronics. Specifically, our experiments exhibit robust detection of 12 weed species with minimal overhead, showcasing better performance than state-of-the-art. These improvements align with sustainable agricultural methods, ensuring that the method remains efficient and suitable given the energy constraints of contemporary farming. Our assessments using the CottonWeedDet12 dataset indicate that our model attains performance on par with larger state-of-the-art (SOTA) architectures, excelling in accuracy, precision, recall, and mean average Precision, yet at lower computational costs. Additionally, we demonstrate the feasibility of deploying EcoWeedNet on devices with limited resources for real-world deployment. This establishes our network as an optimal choice for real-time applications and resource-limited scenarios, making it a significant asset for the progression of next-generation sustainable agricultural technology.

\section*{Acknowledgement}
The authors acknowledge the support received from the Deanship of Research, King Fahd University of Petroleum \& Minerals (KFUPM), and SDAIA-KFUPM Joint Research Center for Artificial Intelligence through Grant\# JRC-AI-RFP-17

\bibliographystyle{IEEEtran}
\bibliography{references}

\end{document}